\documentclass[10pt,twocolumn,letterpaper]{article}

\usepackage[pagenumbers]{cvpr} 

\usepackage[accsupp]{axessibility}  
\usepackage[ruled,lined,linesnumbered]{algorithm2e}
\usepackage{setspace}

\usepackage{graphicx}
\usepackage{amsmath}
\usepackage{amssymb}
\usepackage{booktabs}
\usepackage{mathrsfs}
\usepackage{bm}
\usepackage{multirow}
\usepackage{cite}
\usepackage{times}
\usepackage{epsfig}
\usepackage{booktabs}
\usepackage{bbding}
\usepackage[noend]{algpseudocode}
\usepackage{caption,subcaption}

\usepackage{footnote}
\makesavenoteenv{tabular}
\makesavenoteenv{table}
\makesavenoteenv{figure}
\usepackage[dvipsnames]{xcolor}
\usepackage{pythonhighlight}

\usepackage{xcolor,colortbl}
\definecolor{Gray}{gray}{0.90}
\newcolumntype{g}{>{\columncolor{Gray}}c}
\definecolor{ffe1da}{RGB}{255,225,218}
\definecolor{F7E0D5}{RGB}{247,224,213}
\definecolor{darkF7E0D5}{RGB}{209,154,128}
\colorlet{Light}{White!0!F7E0D5}

\usepackage[pagebackref,breaklinks,colorlinks]{hyperref}
\usepackage{pifont}
\newcommand{\xmark}{\ding{55}}%

\newcommand{\rownumber}[1]{\textcolor{darkF7E0D5}{#1}}

\makeatletter
\newcommand{\removelatexerror}{\let\@latex@error\@gobble}
\makeatother

\usepackage[capitalize]{cleveref}
\crefname{section}{Sec.}{Secs.}
\Crefname{section}{Section}{Sections}
\Crefname{table}{Table}{Tables}
\crefname{table}{Tab.}{Tabs.}

\def\cvprPaperID{2493} 
\def\confName{CVPR}
\def\confYear{2022}

\begin{document}

\title{Towards Discriminative Representation: \\ 
Multi-view Trajectory Contrastive Learning for Online Multi-object Tracking}

\author{En Yu$^1$\thanks{Equal contribution}~, Zhuoling Li$^{2*}$, Shoudong Han$^1$\thanks{Corresponding author}\;\\
$^1$Huazhong Univerisity of Science and Technology\quad $^2$Tsinghua University\\
{\tt\small \{yuen, shoudonghan\}@hust.edu.cn \quad lzl20@mails.tsinghua.edu.cn}
}
\maketitle

\begin{abstract}
Discriminative representation is crucial for the association step in multi-object tracking. Recent work mainly utilizes features in single or neighboring frames for constructing metric loss and empowering networks to extract representation of targets. Although this strategy is effective, it fails to fully exploit the information contained in a whole trajectory. To this end, we propose a strategy, namely multi-view trajectory contrastive learning, in which each trajectory is represented as a center vector. By maintaining all the vectors in a dynamically updated memory bank, a trajectory-level contrastive loss is devised to explore the inter-frame information in the whole trajectories. Besides, in this strategy, each target is represented as multiple adaptively selected keypoints rather than a pre-defined anchor or center. This design allows the network to generate richer representation from multiple views of the same target, which can better characterize occluded objects. Additionally, in the inference stage, a similarity-guided feature fusion strategy is developed for further boosting the quality of the trajectory representation. Extensive experiments have been conducted on MOTChallenge to verify the effectiveness of the proposed techniques. The experimental results indicate that our method has surpassed preceding trackers and established new state-of-the-art performance.
\end{abstract}
\vspace{-5mm}

\section{Introduction}
\label{sec:intro}

\begin{figure}[ht]
\centering
\includegraphics[width=1.0\columnwidth]{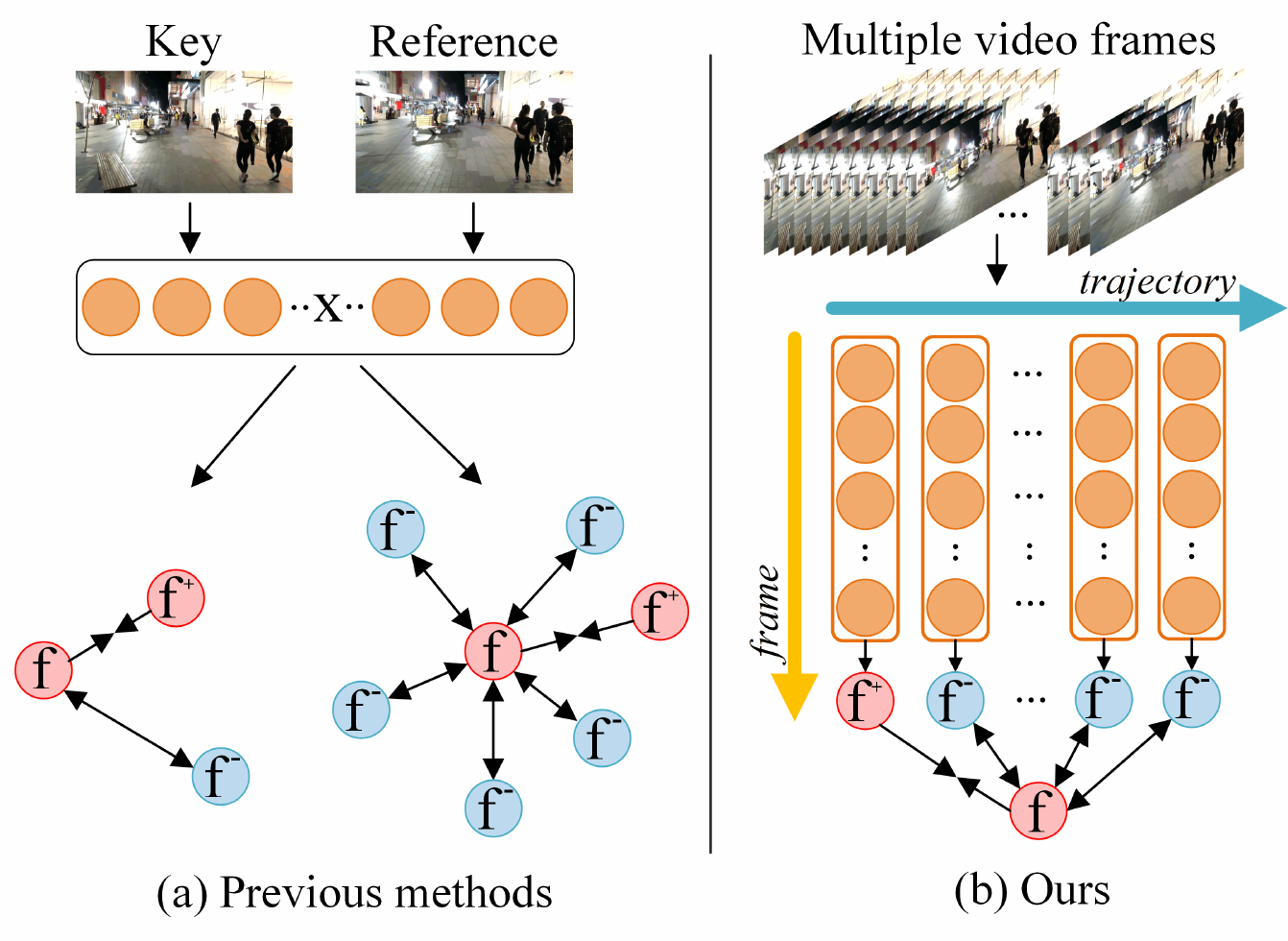}
\caption{\textbf{Comparison between existing methods and our proposed method.} (a) Existing methods only utilize the information in a single or two adjacent frames to learn representation. (b) On the contrary, our method fully exploits the features in the whole trajectories, which contain numerous frames.}
\label{Fig.1} \label{Idea of MTCL}
\vspace{-2mm}
\end{figure}

As a fundamental vision perception task, multiple object tracking (MOT) has been extensively deployed in broad applications, e.g., autonomous driving, video analysis and intelligent robots \cite{bewley2016simple,wojke2017simple}. Previous MOT methods mainly adopt the tracking-by-detection paradigm \cite{tracktor, peng2020chained, Pang2020CVPR, wang2019towards}, which mainly comprises two parts, i.e, detection and association. For the detection part, a detector is established to localize objects of interest. In the association part, some methods utilize a motion predictor for forecasting the positions of objects in the next frames and rely on the position information to associate them \cite{han2020mat}. However, when these methods are applied to the challenging cases where targets are missing for several frames, it is hard to reconnect these targets to the corresponding trajectories correctly. 

To alleviate this problem, existing trackers seek help from appearance-based strategies \cite{wang2019towards, zhang2021fairmot, yu2021relationtrack}, in which objects are identified based on the similarity of extracted features. Nevertheless, the effectiveness of the appearance-based association strategy is still limited. Many objects with different identities are associated with the same trajectory because they are occluded or blurry, thus causing the learned representation to be indistinguishable. Hence, extracting more meaningful and discriminative representation is desired for enhancing the association accuracy. 

\begin{figure}[t]
\centering
\includegraphics[width=1.0\columnwidth]{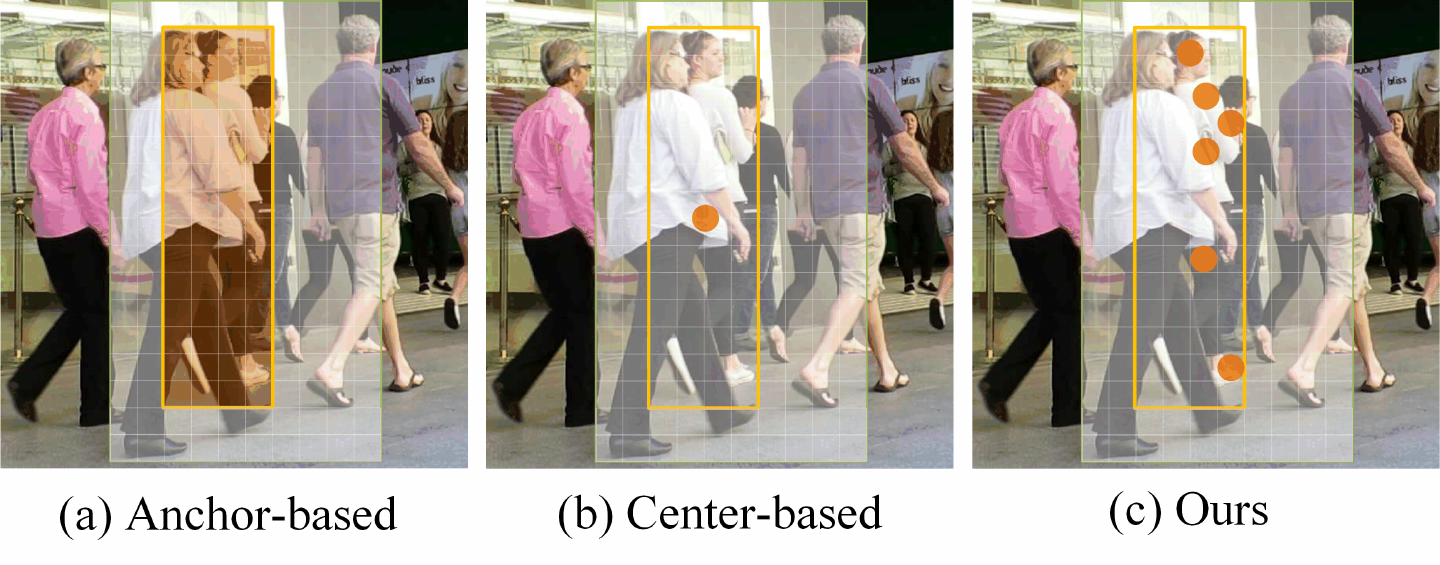} 
\caption{\textbf{Comparison among various strategies for representating targets}: (a) Anchor-based, (b) Center-based, (c) Learnable view sampling (ours).} \label{representating target strategies}
\vspace{-4mm}
\end{figure}  

In order to improve the quality of the extracted representation, we revisit existing representation learning methods in MOT and observe that they only use samples in a single or neighboring frames to construct loss for training networks \cite{wojke2017simple, he2021learnable, wang2019towards, zhang2021fairmot, pang2021quasi}, which is illustrated in Fig.~\ref{Idea of MTCL} (a). However, an object usually appears in many frames of a video, which compose a trajectory. Almost all existing methods fail to make full use of the information contained in the whole trajectories. Given this observation, we raise a new question:  \emph{is it feasible to fully exploit the trajectory information for boosting the discriminability of the target representation?}

A possible solution to this question is constructing contrastive loss \cite{he2020momentum} using all target representation vectors in trajectories. Nevertheless, since the trajectories in a video could include thousands of instances, this solution requires massive computing resource, which is unaffordable. To address this problem, we propose a strategy named \textit{multi-view trajectory contrastive learning} (MTCL). In this strategy, we first model every trajectory as a vector, namely \textit{trajectory center}, and establish a trajectory-center memory bank (TMB) to maintain these trajectory centers.  Every trajectory center in this memory bank is updated dynamically during the training process. Afterwards, we regard the target appearance vectors as queries and devise a contrastive loss to draw them closer to their corresponding trajectory centers while farther away from other trajectory centers, which is shown in Fig.~\ref{Idea of MTCL} (b). In this way, our method is able to exploit the inter-frame trajectory information while only consuming limited memory.

Moreover, we develop a strategy named \textit{learnable view sampling} (LVS), which serves as a subcomponent of MTCL to explore the intra-frame features. As depicted in Fig.~\ref{representating target strategies}, LVS represents each target with multiple adaptively selected keypoints rather than anchors or their 2D centers. These keypoints gather at the meaningful locations of the targets and provide richer views to the aforementioned trajectory contrastive learning process. Additionally, LVS has an extra benefit. As illustrated in Fig.~\ref{representating target strategies} (a) and (b), the anchors or 2D centers of targets are occluded by other objects, while LVS can still focus on visible regions adaptively.       

Furthermore, in the inference stage, we note that the target features of some frames are unclear and inappropriate to represent trajectories. Correspondingly, we devise a \textit{similarity-guided feature fusion} (SGFF) strategy that adaptively aggregates features based on the historical feature similarity to alleviate the influence of these poor features on the trajectory representation.

Incorporating all the above proposed techniques, the resulting model, namely \textit{multi-view tracker} (MTrack), has been evaluated on four public benchmarks, i.e., MOT15 \cite{leal2015motchallenge}, MOT16 \cite{milan2016mot16}, MOT17 \cite{milan2016mot16} and MOT20 \cite{dendorfer2020mot20}. The experimental results indicate that all our proposed strategies are effective and MTrack outperforms preceding counterparts significantly. For instance, MTrack achieves IDF1 of $69.2\%$ and MOTA of $63.5\%$ on MOT20. 

\section{Related Work}

\noindent \textbf{Multiple-object Tracking.} Thanks to the rapid development of 2D object detection techniques \cite{ren2015faster, tian2019fcos, zhou2019objects, carion2020end}, recent trackers mainly adopt the tracking-by-detection paradigm \cite{bewley2016simple, wojke2017simple, tracktor, peng2020chained, Pang2020CVPR, wang2019towards}. The trackers following this paradigm first utilize detectors to localize targets in each frame and then employ an associator to link detected objects of the same identity to form trajectories. 

Traditional methods \cite{bewley2016simple} usually perform temporal association via motion-based algorithms, such as Kalman Filter \cite{welch1995introduction} and optical flow \cite{baker2004lucas}. However, these algorithms behave poorly when targets move irregularly. In contrast to these traditional methods, some recent methods employ neural networks to predict the locations and displacements of targets in the next frames jointly \cite{feichtenhofer2017detect, tracktor, CenterTrack, sun2020transtrack, shuai2021siammot}. However, when these methods are applied to complex scenarios where objects are occluded and invisible for several frames, the tracking results become unsatisfactory.

To mitigate the aforementioned problems, appearance-based methods \cite{wojke2017simple, he2021learnable, wang2019towards, zhang2021fairmot, liang2020rethinking, wu2021track, yu2021relationtrack} are introduced. These methods utilize neural networks to extract features of detected targets. The extracted features are desired to be discriminative, which means the features of objects with the same identity are similar while the ones corresponding to different identities are diverse. Since the appearance-based methods associate targets based on the similarity of extracted features, how to produce discriminative features is critical for them. In this work, we propose the MTCL which enables a model to learn more discriminative features through the proposed trajectory contrastive learning.

\begin{figure*}[t]
\centering
\includegraphics[width=1.7\columnwidth]{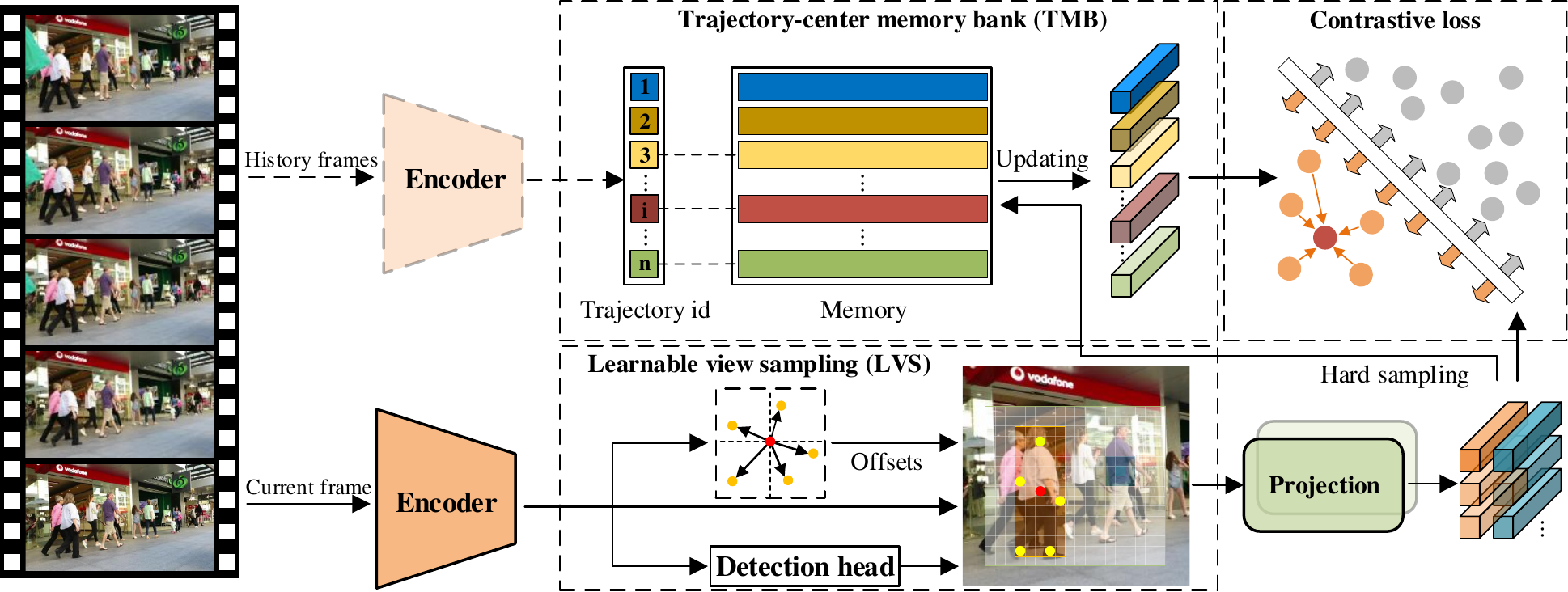} 
\caption{\textbf{Overall pipeline of multi-view trajectory contrastive learning}. Given a video $\mathbb{V}_{\xi}$ with $\xi$ frames $I_{t}$ (t=1,2,...,$\xi$), MTCL comprises 4 steps: (1) Employ an encoder (the backbone) to extract feature maps from the current input frame. (2) Use LVS to select informative keypoints from the extracted feature maps, and transform the features of selected keypoints as target appearance vectors by a projection head. (3) Conduct contrastive learning between the appearance vectors and trajectory centers stored in the memory bank. (4) Update trajectory-center memory bank using our hard sampling strategy.}
\label{meth_mtcl}
\vspace{-3.5mm}
\end{figure*}

\vspace{1mm}
\noindent \textbf{Constrastive Learning.} Contrastive learning has been widely studied due to its impressive performance in the field of self-supervised learning \cite{chen2020simple, he2020momentum, hu2021adco, wang2021dense}. Given some images, contrastive learning first transforms every image into various views through random augmentation. Its optimization objective is drawing a view closer to the views augmented from the same image, and pushing it away from the views that originate from other images \cite{he2020momentum}.

Although contrastive learning has been deployed in many fields, such as classification \cite{chen2020improved,caron2020unsupervised} and detection \cite{xie2021detco,yang2021instance}, few works applies it to MOT. Recently, QDTrack \cite{pang2021quasi} serves as the first work that utilizes contrastive learning in MOT to improve the learned representation. However, it only uses the samples in adjacent two frames and fails to explore information in the whole trajectories. This is because directly constructing contrastive loss with all samples in trajectories leads to a tremendous computing burden, which is unaffordable. On the contrary, our proposed MTCL can fully exploit the trajectory information with very limited computing resource, which is realized by only storing one vector for every trajectory in the memory.

\section{Methodology}
This section explains how MTrack is implemented. First of all, Sec.~\ref{overview} presents an overview of MTrack. Afterwards, Sec.~\ref{MTCL} introduces our proposed MTCL, which includes two subcomponents, i.e., LVS and TMB. Finally, Sec.~\ref{similarity_guided} describes the implementation of the SGFF strategy.

\subsection{Overview} 
\label{overview}

In this work, we implement MTrack based on CenterNet \cite{zhou2019objects}, which represents targets as their center points. DLA34 \cite{yu2018deep} is adopted as the backbone of the CenterNet. Given a video $V_{\xi}$ of $\xi$ frames as input, the backbone is firstly applied to extract feature maps. Then, several network heads are established to transform the feature maps into the desired properties of the targets, which include 2D center heatmaps, center offsets and bounding box sizes. Besides, we add an extra embedding head in parallel with the detection heads to extract appearance features. In our implementation, the detection bounding boxes are generated based on the estimated 2D center heatmaps, center offsets and bounding box sizes. The detected objects are associated according to their appearance feature similarities.

In this work, we focus on how to produce discriminative representation for realizing accurate association. Specifically, MTCL is applied during the training process to improve the ability of generating representative embedding vectors, which is illustrated in Fig.~\ref{meth_mtcl}. SGFF is developed for improving the quality of the trajectory representation during the inference phase.

\subsection{Multi-view Trajectory Contrastive Learning}
\label{MTCL}

In this subsection, we elaborate on our main contribution, MTCL. To explain it clearly, we first introduce the LVS strategy in MTCL, which generates multiple appearance vectors adaptively for every target and contributes to exploiting the intra-frame information more efficiently. Then, we describe the trajectory-center memory bank, which enables us to realize trajectory contrastive learning with only limited computing resource. Finally, we present the details of the trajectory-level contrastive loss and the overall process of MTCL.
 
\vspace{1mm}
\noindent \textbf{Learnable view sampling.} Existing trackers developed based on CenterNet mainly represent every target as a sole center point on feature maps. As introduced before, this strategy has two critical restrictions: (1) The center points of targets could be occluded by other objects, such as the case shown in Fig.~\ref{representating target strategies} (b). In this case, the produced appearance vectors fail to reflect the characteristics of targets. (2) Representating every target with only one vector cannot provide sufficient samples to the contrastive learning algorithm.

To address the above restrictions, we devise LVS that represents a target as multiple adaptively selected keypoints. Specifically, given the $t_{\rm th}$ frame of a video as input, denoted as $I_{\rm t}$, we first transform it into feature maps $F_{\rm t}$ using the backbone, and recognize the center points of all targets in $I_{\rm t}$ like former trackers \cite{zhang2021fairmot}. Denoting the center point coordinate of a target $q \in I_{t}$ on $F_{\rm t}$ as $Z^{q} = (x^{q}, y^{q})$, we take out the vector at the coordinate of $Z^{q}$ in $F_{\rm t}$ and represent this vector as $r^{q}$. The vector $r^{q}$ contains the appearance information of $q$. Therefore, we can regress the offsets from $Z^{q}$ to the potential informative keypoints in $I_{\rm t}$ by applying a linear transformation $W$. This process can be formulated as 
\begin{align}
\triangle Z^{q} = W r^{q}, \label{Eq1}
\end{align}
where $\triangle Z^{q} = \{\triangle Z^{q}_{i}\}_{i=1}^{N_{k}}$, $N_{k}$ represents the total number of selected keypoints and $\triangle Z^{q}_{i}$ is the offset from $Z^{q}$ to the $i_{\rm th}$ keypoint. Hence, the coordinate of the $i_{\rm th}$ selected keypoint is determined as:
\begin{align}
Z^{k}_{i} = \Phi (Z^{q} + \triangle Z^{q}_{i}), \label{Eq2}
\end{align}
where $\Phi(\cdot)$ denotes an operator that guarantees all the keypoints to fall within the generated 2D bounding boxes. Specifically, if a keypoint is out of a 2D box, it will be clipped to the boundary of this box.

With the coordinate of the $i_{\rm th}$ keypoint $Z^{k}_{i}$, we can obtain its corresponding feature vector $v^{k}_{i}$ from $F_{\rm t}$. Then, $v^{k}_{i}$ is further transformed into a more representative appearance vector $\widetilde{v}^{k}_{i}$  by applying a projection head consisting of 4 fully connected layers. Since $N_{k}$ keypoints are selected for every target, we can obtain $N_{k}$ appearance vectors $\widetilde{v}^{k}_{i}$ corresponding to different positions of a target. Representing every detected target as these $N_{k}$ vectors provides richer intra-frame samples to the contrastive learning process. Notably, during the inference stage, the $N_{k}$ appearance vectors are contatenated as a single one to represent their corresponding target.

\vspace{1mm}
\noindent \textbf{Trajectory-center memory bank.} Compared with intra-frame samples, inter-frame samples provide more informative features. However, directly utilizing all samples in the historical frames to construct contrastive loss consumes much computing resource. To alleviate this problem, we propose to represent every trajectory as a vector named \textit{trajectory center}, and maintain all the trajectory centers in a memory bank. 

Assuming there are $N$ trajectories in a training dataset, the memory bank is initialized as a set containing $N$ zero vectors $\{c_{i}\}_{i=1}^{N}$ at the beginning of a training epoch (a training epoch includes many iterations, and every iteration corresponds to an input data batch). This memory bank and its contained trajectory centers are not reinitialized until the end of this epoch. Since we do not save the gradient information of all iterations for saving memory, the trajectory centers cannot be updated by gradient-based optimizers, such as Adam \cite{kingma2014adam}. To solve this problem, we develop a momentum-based updating strategy that dynamically updates trajectory centers in each iteration without requiring historical gradient information.

Hereby, we explain how the momentum-based updating strategy is implemented. For an input data batch comprising several frames, the network recognizes the instances of interest in all frames and generates $N_{k}$ appearance vectors for each instance. During the training stage, every detected instance is labeled with a trajectory ID. We gather all appearance vectors extracted from the instances with the same trajectory ID to update their corresponding trajectory centers.

Specifically, the trajectory centers are updated after the parameters of the model are optimized based on the back-propagated gradient.  Denoting all appearance vectors in a batch corresponding to the $l_{\rm th}$ trajectory as ${\rm P}^{l}=\{p_{i}^{l}\}_{i=1}^{N_{l}}$, we select the hardest sample in ${\rm P}^{l}$ to update $c_{l}$. The hard levels of samples are reflected by the cosine similarities \cite{wang2018cosface} with their corresponding trajectory centers, and the sample with the minimum similarity is the hardest one. Mathematically, the cosine similarity $s_{i}^{l}$ between $p_{i}^{l}$ and $c_{l}$ is formulated as:
\begin{align}
s_{i}^{l} = \frac{p_{i}^{l} \cdot c_{l}}{\Vert p_{i}^{l}\Vert_2 \times \Vert c_{l} \Vert_2}, \label{Eq3}
\end{align}
where $\cdot$ and $\times$ represent the dot product between two vectors and the normal product between two floating numbers, respectively. $\Vert \cdot \Vert_2$ is a $L_{2}$ normalization operator.

Denoting the appearance vector in ${\rm P}^{l}$ with the minimum cosine similarity as $p_{m}^{l}$, $c_{l}$ is updated given:
\begin{align}
c_{l} \leftarrow \alpha c_{l} + (1-\alpha) p_{m}^{l}, \label{Eq4}
\end{align}
where $\alpha$ is a hyper-parameter falling between [0, 1].

During the training phase, updating trajectory centers with hard samples contributes to the efficiency of training a network. This issue has been confirmed by our experimental results.

\vspace{1mm}
\noindent \textbf{Trajectory-level contrastive loss.} LVS and TMB provide rich intra-frame and inter-frame samples for constructing the contrastive loss, respectively. The following problem is how to implement the contrastive loss, and train the embedding head of the network for producing discriminative representation.

For the $k_{\rm th}$ appearance vector $\tilde{v}^{k}_{l}$ produced by LVS and its corresponding trajectory center $c_{l}$, the optimization objective is drawing  $\tilde{v}^{k}_{l}$ closer to $c_{l}$ while pushing $\tilde{v}^{k}_{l}$ away from all other trajectory centers. Following \cite{he2020momentum}, we employ the InfoNCE loss to realize this objective. Mathematically, the loss of $\tilde{v}^{k}_{l}$ can be formulated as:

\begin{align}
L_{NCE}^{k} = -log \frac{ exp(\tilde{v}^{k}_{l} \cdot c_{l}) / \tau }{\sum_{i=0}^{N_{t}}exp(\tilde{v}^{k}_{l} \cdot c_{i}) / \tau}, \label{Eq5}
\end{align}
where $\tau \in (0,1] $ denotes a hyper-paramter and $N_{t}$ is the total number of the trajectories. To fully exploit inter-frame information, we calculate the trajectory-level contrastive loss for every appearance vector based on Eq.(\ref{Eq5}), and the embedding head loss $L_{tcl}$ is formulated as: 
\begin{align}
L_{tcl} = \frac{1}{N_{a}} \sum\limits_{k=1}^{N_{a}} L_{NCE}^{k}, \label{Eq5_2}
\end{align}
where $N_{a}$ is the total number of appearance vectors. Overally, the total loss for training MTarck is:
\begin{align}
L = \frac{1}{2}(\frac{1}{e^{\eta_{1}}}L_{det}+\frac{1}{e^{\eta_{2}}}L_{tcl} + \eta_{1} + \eta_{2}), \label{Eq6}
\end{align}
where $L_{det}$ represents the detection loss. $\eta_{1}$ and $\eta_{2}$ are learnable weights for balancing $L_{det}$ and $L_{tcl}$.

Additionally, we give the pseduo code of MTCL in Algorithm~\ref{alg:training} to present its process clearly. 

\vspace{1mm}
\begin{minipage}{0.45\textwidth}
\removelatexerror
\begin{algorithm}[H]
  \caption{Training Procedure of MTCL}
  \label{alg:training}
  \setstretch{1.1}
  \SetAlgoLined
  \SetKwInOut{Require}{Require} \SetKwInOut{Input}{Input}
      \Require{
      The feature encoder $\sigma_{\theta}$;\\
      Momentum rate $\alpha$;\\
      Temperature parameter $\tau$; \\
      } 
      \Input{Training videos $\textbf{V}=\{V_1,V_2,...V_N\}$;}
      \For{$each\ \ epoch$}{
          Initialize a trajectory-center memory bank $\textbf{B}$; \\
          \For{$each\ \ mini\_batch$}{
              Extract feature maps $F_{b}$ by $\sigma_{\theta}(V_{b})$; \\
              Detect all targets in $V_b$ given $F_{b}$; \\
              Generate multiview appearance vectors by LVS with Eq.~(\ref{Eq1}) and Eq.~(\ref{Eq2}); \\
              Compute trajectory-level contrastive loss $L_{tcl}$ with Eq.~(\ref{Eq5}) and Eq.~(\ref{Eq5_2}); \\
              Update $\textbf{B}$ with Eq.~(\ref{Eq4});
          }
      }
\end{algorithm}
\end{minipage}
\vspace{-3mm}


\subsection{Similarity-guided Feature Fusion}
\label{similarity_guided}
 
In the inference phase, existing appearance-based trackers associate targets with trajectories based on the appearance similarities. First of all, these trackers compute a pair-wise appearance affinity matrix between the targets and trajectories. After that, they associate the targets to the trajectories based on a greedy matching strategy, such as the Hungarian algorithm \cite{kuhn1955hungarian}. In this process, the discriminability of the trajectory representation is critical for the association accuracy.

Existing methods update the representation of trajectories by fusing the features of historical frames. Denoting the representation of the $l_{\rm th}$ trajectory in the $(t-1)_{\rm th}$ frame as $f_{l}^{t-1}$, these methods employ a momentum-based updating strategy \cite{zhang2021fairmot} to update $f_{l}^{t-1}$ with the feature vector $z_{l}^{t}$, which is extracted from the newly matched object. Then, $f_{l}^{t}$ is obtained. This process is formulated as follows:
\begin{align}
f_{l}^{t} = (1 - \beta) f_{l}^{t-1} + \beta z_{l}^{t}, \label{eq_update}
\end{align} 
where $\beta$ is a hyper-parameter.  

Setting $\beta$ to a constant is effective when $z_{l}^{t}$ is informative. However, when targets are occluded or blurry, $z_{l}^{t}$ is contaminated by noise and contains little valuable information. Under this condition, updating $f_{l}^{t-1}$ with $z_{l}^{t}$ is harmful to the trajectory representation.  

\begin{figure}[t]
\centering
\includegraphics[width=1.0\columnwidth]{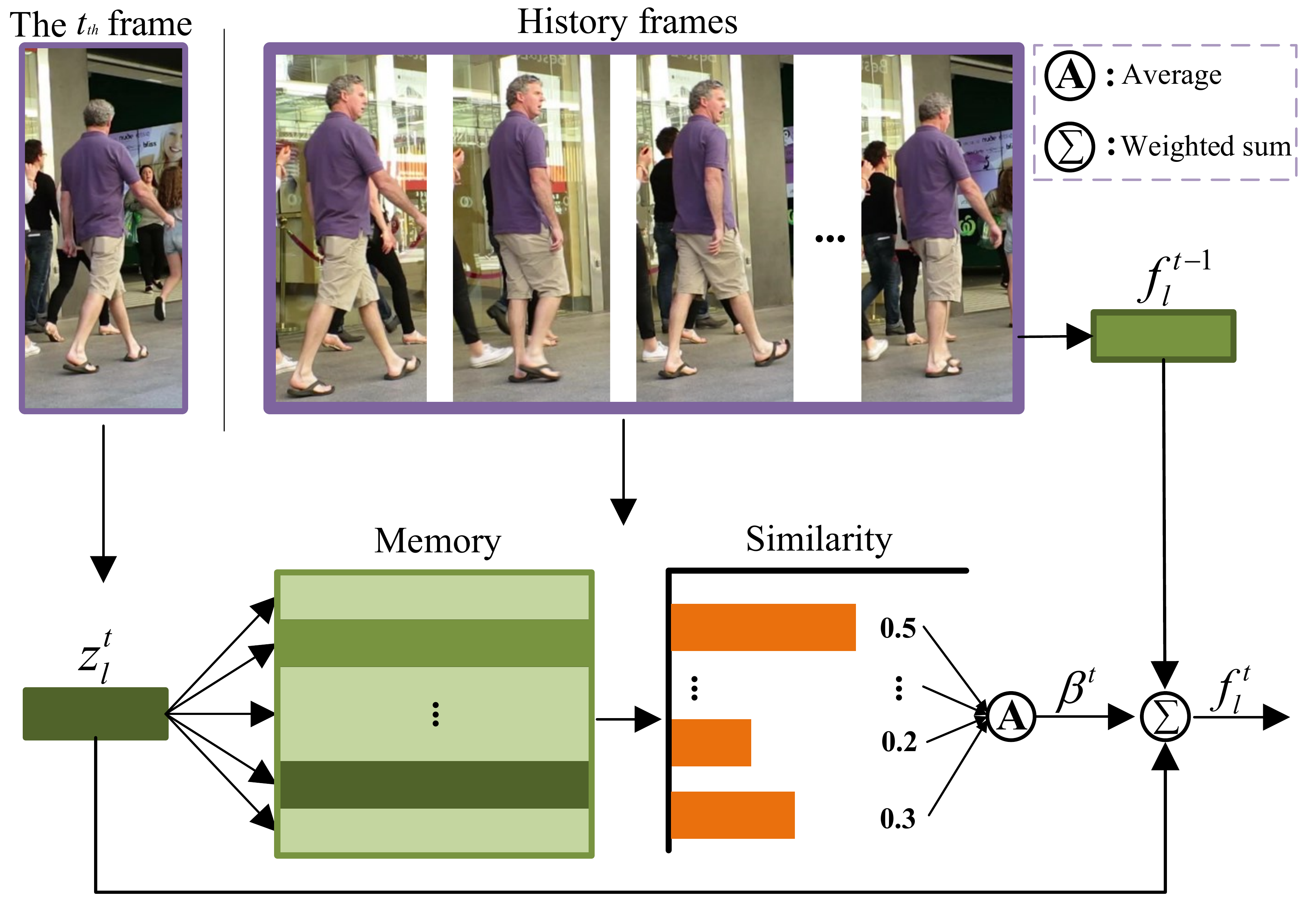} 
\caption{\textbf{Illustration of the similarity-guided feature fusion strategy.} In this strategy, $\beta^{t}$ is adjusted adaptively for each frame according to the similarities between $z_{l}^{t}$ and the features extracted from the recent frames.}
\label{fig:similarity}
\vspace{-5mm}
\end{figure}

To deal with this problem, we propose the SGFF that adjusts $\beta$ for each frame adaptively, which is shown in Fig.~\ref{fig:similarity}. Specifically, assuming a target is clear in most recent frames, we can measure the quality of $z_{l}^{t}$ by computing its similarities with the feature vectors in the latest $Q$ frames. If $z_{l}^{t}$ is similar to them, we can speculate that $z_{l}^{t}$ is informative and $\beta$ should be a large value. Therefore, denoting $\beta$ in the $t_{\rm th}$ frame as $\beta^{t}$, $\beta^{t}$ can be derived as:
\begin{align}
\beta^{t} = \max\{0, \frac{1}{Q} \sum_{i=1}^{Q} \Psi_{d}(z_{l}^{t}, z_{l}^{t-i})\},
\end{align}  
where $\Psi_{d}(\cdot)$ represents an operator that computes the cosine similarity described in Eq. (\ref{Eq3}).

With the SGFF, $\beta^{t}$ becomes a tiny value if $z_{l}^{t}$ is of poor quality. Thus, this strategy reduces the effect of the poor feature vectors.

\section{Experiments}

This section presents the experimental details. Specifically, Sec.~\ref{data and eval} introduces the adopted datasets as well as the evaluation metrics. Sec.~\ref{implementation} describes the implementation details of our method. Then, Sec.~\ref{comparision} demonstrates the superiority of MTrack by comparing it with existing state-of-the-art (SOTA) MOT methods. Sec.~\ref{abalation} reveals the effectiveness of the proposed techniques through various ablation experiments. Finally, Sec.~\ref{vis} visualizes some extracted features and suggests that our method boosts the discriminability of the learned representation significantly.

\subsection{Datasets and Evaluation Metrics}
\label{data and eval}

\noindent \textbf{Datasets:} We conduct extensive experiments on four public MOT benchmarks, i.e., MOT15 \cite{leal2015motchallenge}, MOT16 \cite{milan2016mot16}, MOT17 \cite{milan2016mot16} and MOT20 \cite{dendorfer2020mot20}. Specifically, MOT15 comprises 22 sequences, one half for training and the other half for testing. This dataset in all contains 996 seconds of videos, which includes 11286 frames. MOT16 and MOT17 are composed of the same 14 videos, 7 for training and the other 7 for testing. The 14 videos cover various scenarios, viewpoints, camera poses and weather conditions. Compared with MOT16, MOT17 provides more detection bounding boxes produced by various detectors, which include DPM \cite{felzenszwalb2009object}, SDP \cite{yang2016exploit}, Faster-RCNN \cite{ren2015faster}. MOT20 is the most challenging benchmark in these datasets. It consists of 8 video sequences captured in 3 crowded scenes. In some frames, more than 220 pedestrians are included simultaneously. Meanwhile, the scenes in MOT20 are very diverse, which could be indoor or outdoor, in the day or at night.

\noindent \textbf{Evaluation metrics:} MTrack is evaluated based on the CLEAR-MOT Metrics \cite{bernardin2008evaluating}, which include ID F1 score (IDF1), multiple object tracking accuracy (MOTA), multiple object tracking precision (MOTP), mostly tracker rate (MT), mostly lost rate (ML), false positives (FP), false negatives (FN) and identity switches (IDS). Among them, IDF1 and MOTA are the most important indexes for comparing performance. 

\subsection{Implementation Details}
\label{implementation}

We adopt DLA-34 as the backbone and the detection branch of MTrack is pre-trained on Crowdhuman \cite{shao2018crowdhuman}. The parameters are updated using the Adam optimizer \cite{kingma2014adam} with the initial learning rate of $10^{-4}$. The learning rate is reduced to $10^{-5}$ at the $20_{\rm th}$ epoch. The model is trained for 30 epochs totally. During the training process, the batch size is set as $8$ and the resolution of every input image is $1088 \times 608$. The adopted image augmentation operations follow FairMOT \cite{zhang2021fairmot}, which include random rotation, scaling, translation and color jittering. For LVS, we set $N_{k}$ to 9, and choose the center point and the eight points surrounding it as the initial sampling locations. In TMB, the temperature parameter $\tau$ is $0.05$ and the momentum update factor $\alpha$ is $0.2$. During the inference stage, $Q$ is set as $30$.

\subsection{Comparison with Preceding SOTAs}
\label{comparision}

In this part, we compare the performance of MTrack with preceding SOTA methods on four widely adopted benchmarks, i.e., MOT15, MOT16, MOT17 and MOT20. The results are reported in Tab.~\ref{table:sota}. Notably, some methods utilize numerous extra data with identity labels to improve their abilities on generating discriminative identity embedding. For fair comparison, we do not use extra training data from the other tasks (such as person search or ReID) to boost the tracking performance. According to the results in Tab.~\ref{table:sota}, our method surpasses all the compared counterparts significantly on IDF1, MOTA and the other metrics.

\begin{table}[t]
	\footnotesize
	\begin{center}
		\setlength{\tabcolsep}{0.65mm}{
		\begin{tabular}{lcccccccc}
			\toprule
			 Method & IDF1 & MOTA & MT & ML$\downarrow$ & FP$\downarrow$ & FN$\downarrow$ & IDS$\downarrow$\\
			\midrule
			\multicolumn{8}{@{}l}{\textit{MOT15 private detection}}\\
			DMT\cite{kim2016cdt} & 49.2 & 44.5 & 34.7\% &22.1\% & 8,088 & 25,335 & 684 \\
			
			TubeTK \cite{Pang2020CVPR}& 53.1 & 58.4 & 39.3\%& 18.0\%& \textbf{5,756}& 18,961 &854\\
			 
			CDADDAL\cite{bae2018confidence} & 54.1 & 51.3 & 36.3\% &22.2\% & 7,110 & 22,271 & 544 \\
			
 			TRID \cite{manen2017pathtrack}& 61.0& 55.7 & 40.6\%& 25.8\%& 6,273 & 20,611 & \bf 351 \\
			 
			RAR15 \cite{fang2018recurrent}& 61.3 & 56.5 & \bf 45.1\%& \bf 14.6\%& 9,386& \bf 16,921 &428\\ \rowcolor{gray!40}
			 

			\rowcolor{Light} \textbf{MTrack} &\bf 62.1 & \bf 58.9& 38.1\% & 17.5\% & 6,314 & 18,177 & 750 \\
			\midrule
			\midrule
			\multicolumn{8}{@{}l}{\textit{MOT16 private detection}}\\
			IoU\cite{bochinski2017high} & 46.9 & 57.1 & 23.6\% &32.9\% &\textbf{5,702}& 70,278 &2,167 \\

			JDE \cite{wang2019towards} & 55.8 & 64.4 & 35.4\% & 20.0\% & - & - & 1544 \\
			
			CTracker\cite{peng2020chained} & 57.2 & 67.6 & 32.9\% &23.1\% &8,934 &48,305 &1,897 \\

			TubeTK \cite{Pang2020CVPR}& 59.4& 64.0& 33.5\%& 19.4\%& 10,962 &53,626 &1,117\\
			
			LMCNN \cite{babaee2019dual}& 61.2& 67.4 & 38.2\%& 19.2\%& 10,109 &48,435 &931\\
			
			DeepSort \cite{wojke2017simple} & 62.2 & 61.4 & 32.8\%& 18.2\% & 12,852 & 56,668 & 781\\

			MAT \cite{han2020mat}& 63.8& 70.5& 44.7\%& 17.3\% & 11,318 & 41,592 & 928\\ 
			
			TraDeS \cite{wu2021track} & 64.7 & 70.1 & 37.3\% & 20.0\% & 8,091 & 45,210 & 1,144\\

			MOTR \cite{zeng2021motr} & 67.0& 65.7&  37.2\%& 20.9\%& 16,512& 45,340& 648 \\

			QDTrack \cite{pang2021quasi} & 67.1 & 69.8 & 41.6\% &  19.8\% & 9,861, & 44,050 & 1,097 \\
			

			GMTCT \cite{he2021learnable} & 70.6 & 66.2 & 29.6\% & 30.4\% & 6,355 & 54,560 & 701 \\

\rowcolor{Light}
			\textbf{MTrack} &\textbf{74.3}&\textbf{72.9}&\textbf{50.6\%} &\bf 15.7\% &19,236  &\textbf{29,554} & \textbf{642}\\
			\midrule
			\midrule
			\multicolumn{8}{@{}l}{\textit{MOT17 private detection}}\\
			DAN \cite{sun2019deep} & 49.5 & 52.4 & 21.4\% &30.7\% &25,423 &234,592 &8,431 \\
		

			Tracktor+CTdet \cite{tracktor} & 57.2 & 56.1 & 25.7\% & 29.8\% & 44,109 & 210,774 & 2,574 \\

			CTracker \cite{peng2020chained} & 57.4 & 66.6 & 32.2\% &24.2\% &22,284 &160,491 & 5,529 \\
			 
			TubeTK \cite{Pang2020CVPR}& 58.6& 63.0& 31.2\%& 19.9\%& 27,060 &177,483 &5,727\\

			TransCener \cite{xu2021transcenter} & 62.1 & 70.0 & 38.9\% & 20.4\% & 28,119 & 136,722 & 4,647 \\

			MAT \cite{han2020mat}& 63.1& 69.5& 43.8\%& 18.9\%& 30,660& 138,741& 2,844\\ 

			TraDeS \cite{wu2021track} & 63.9 & 69.1 & 37.3\% & 20.0\% & 20,892 & 150,060 & 3,555\\

			CenterTrack \cite{CenterTrack} & 64.7 & 67.8 & 34.6\% & 24.6\% & 18,489 & 160,332 & 3,039 \\

			MOTR \cite{zeng2021motr} & 66.4& 65.1& 33.0\%& 25.2\%& 45,486& 149,307& 2,049 \\  

			GMTCT \cite{he2021learnable} & 68.7 & 65.0 & 29.4\% & 31.6\% & \bf 18,213 & 177,058 & 2,200 \\

			QDTrack \cite{pang2021quasi} & 68.7 & 66.3 & 40.6\% & 21.9\% & 26,589, & 146,643 & 3,378 \\
			

			
\rowcolor{Light}			
			\textbf{MTrack} & \bf 73.5 & \bf 72.1 & \bf 49.0\% & \bf 16.8\% &53,361 & \bf 101,844 & \bf 2,028\\
			\midrule
			\midrule
			\multicolumn{8}{@{}l}{\textit{MOT20 private detection}}\\
			MLT \cite{zhang2020multiplex} & 54.6 & 48.9 & 30.9\% &22.1\% & \bf 45,660 & 216,803 & \bf 2,187 \\

			TransCener \cite{xu2021transcenter} & 50.4 & 61.9 & 49.4\% & 15.5\% & 45,895 & 146,347 & 4,653 \\
			
			FairMOT \cite{zhang2021fairmot}& 67.3 & 61.8 & 68.8\%& 7.6\%& 103,440 & 88,901 & 5,243\\



			\rowcolor{Light} \textbf{MTrack} &\bf 69.2 & \bf 63.5&\bf 68.8\% & \bf 7.5\% &96,123  &\bf 86,964 & 6,031\\
			\bottomrule
		\end{tabular}}
	\end{center}
	\vspace{-4mm}
	\caption{Performance comparison with preceding SOTAs on the testing splits of the MOT15, MOT16, MOT17 and MOT20 benchmarks under the private detection protocols. The best results are marked in bold and our method is highlighted in \colorbox{Light}{pink}.}
	\label{table:sota}
\end{table} 

\noindent \textbf{MOT15:} According to Tab.~\ref{table:sota}, MTrack obtains the metric IDF1 of $62.1\%$ and MOTA of $58.9\%$, which significantly outperforms all compared methods without using extra training data. Despite MOT15 containing many false annotations, MTrack still achieves outstanding performance. 

\noindent \textbf{MOT16 and MOT17:} Tab.~\ref{table:sota} shows our main results on MOT16 and MOT17. Since MOT16 and MOT17 contain more data and the annotations are more precise compared with MOT15, MTrack outperforms the compared methods by larger margins. For instance, MTrack surpasses the tracker CenterTrack, which is also built upon CenterNet, by $8.8 \%(73.5\%-64.7\%)$ on IDF1 and $4.3 \%(72.1\%-67.8\%)$ on MOTA in the MOT17 benchmark. Compared with QDTrack, a method that also applies contrastive learning to MOT, MTrack surpasses it by $4.8\% (73.5\%-68.7\%)$ on IDF1 and $5.8\% (72.1\%-66.3\%)$ on MOTA. The results demonstrate that learning from the entire trajectories in videos is more likely to empower the model to learn discriminative representation than learning from neighboring frames. Moreover, it can be observed that MTrack also behaves well on the metric of FN and IDS, which means the the generated trajectories are very continuous. 

\noindent \textbf{MOT20:} To further prove the effectiveness of our method, we evaluate MTrack on the challenging MOT20 benchmark. As shown in Tab.~\ref{table:sota}, MTrack obtains the metric IDF1 of $69.2\%$ and MOTA of $63.5\%$. It behaves the best compared with the counterparts that do not employ extra training data. Notably, MTrack performs better than FairMOT, which is pre-trained on numerous external training datasets, which include ETH, CityPerson, CalTech, CUHK-SYSU and PRW. MTrack surpasses FairMOT by $1.9\% (69.2\%-67.3\%)$ on IDF1 and $1.7\% (63.5\%-61.8\%)$ on MOTA. The results further confirm the superiority of Mtrack, especially in very crowded scenarios.

\subsection{Ablation Study}
\label{abalation}
In this subsection, we verify the effectiveness of the proposed strategies separately through ablation studies. All the experiments are conducted on the MOT17 dataset. Since MOT Challenge does not provide the validation set, we divide the MOT17 datasets into two parts, $\frac{3}{4}$ as the training set and the other $\frac{1}{4}$ as the validation set. All the models are trained for 30 epochs on the training set of MOT17.

\vspace{1mm}
\noindent \textbf{Analysis of the components of MTrack.} In this part, we verify the effectiveness of various components in MTrack through an ablation study. The results are reported in Tab.~\ref{tab:components}. According to the results, all the components have boosted the tracking performance effectively. Incorporating all of them, MTrack (row \rownumber{5})  outperforms the baseline (row \rownumber{2}) by $3.1\%$ on IDF1 and $2.2\%$ on MOTA. Among these components, TMB (row \rownumber{4}) boosts the results with the largest margin. Specifically, it improves IDF1 by 1.1\% and MOTA by 0.8\%. This issue suggests that taking all the information in the whole trajectories into consideration is valuable for the tracking precision.

\begin{table}[htbp]
  \centering
\small
  \setlength{\tabcolsep}{3.5pt}
  \begin{tabular}{@{}p{.8em}@{}lc ccccc c cc@{}}
    \toprule
     &&& \multicolumn{5}{c}{$Components$} && \multicolumn{2}{c}{$Metrics$} \\
    \cmidrule{4-8}
    \cmidrule{10-11}
	  & Method && LVS & Proj. & TMB & Loss & SGFF && IDF1 & MOTA \\
    \midrule
	  \rownumber{1}&	  Base && \xmark & \xmark & \xmark & CE & \xmark && 78.4 & 71.6 \\
	  \rownumber{2}&	   && \Checkmark & \xmark & \xmark & CE & \xmark && 79.1 & 72.3 \\
	  \rownumber{3}&	   && \Checkmark & \Checkmark & \xmark & CE & \xmark && 79.9 & 72.4\\
	  \rownumber{4}&	   && \Checkmark & \Checkmark & \Checkmark & TCL & \xmark && 81.0 & 73.2 \\
	  \rowcolor{Light} \rownumber{5}&	   MTrack && \Checkmark & \Checkmark &  \Checkmark & TCL & \Checkmark && \bf 81.5 & \bf 73.8 \\
    \bottomrule
  \end{tabular}
  \vspace{-1mm}
  \caption{\textbf{Ablation study of the components in MTrack.} The resulting MTrack that combines all the components is highlighted in \colorbox{Light}{pink} (LVS: learnbale view sampling, TMB: trajectory-center memory bank, Proj.: projection, CE: cross-entropy, TCL: trajectory-level contrastive loss, SGFF: similarity-guided feature fusion).}
\label{tab:components}
\vspace{-4mm}
\end{table} 

According to row \rownumber{3}, adding an extra projection head enhances the result on IDF1 significantly, which is a metric mainly reflecting the association accuracy. This observation is consistent with the conclusion drawn by previous publications about contrastive learning \cite{he2020momentum}. In addition, it can be noticed that the SGFF (row \rownumber{5}) also enhances the results notably (0.5\% on IDF1 and 0.6\% on MOTA), although it does not demand any modification to the training process.

\begin{figure}[htbp]
\centering
\includegraphics[width=0.9\columnwidth]{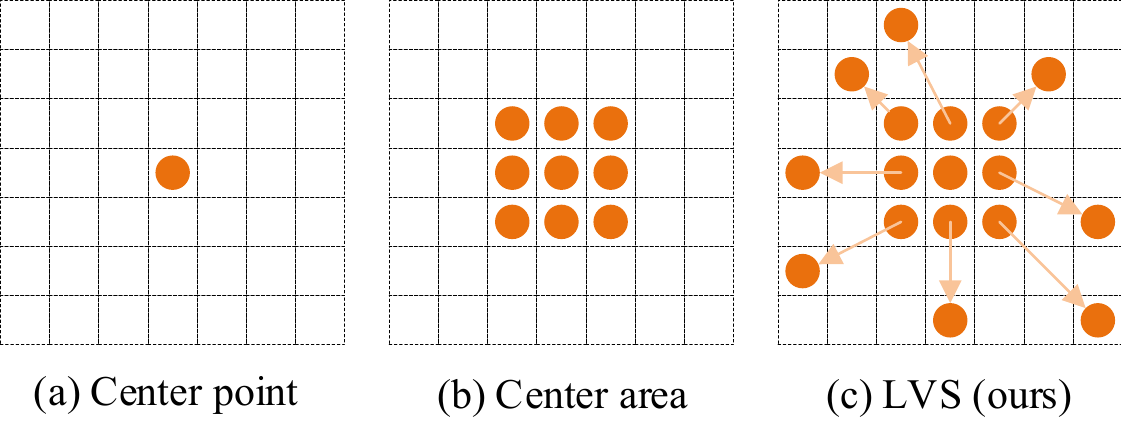}
\caption{Illustration of three positive sampling strategies. (LVS: learnable view sampling)}
\label{exp:lvs}
\end{figure}
\vspace{-2mm}
\vspace{-2mm}
\begin{table}[htbp]
  \centering
\small
  \setlength{\tabcolsep}{3pt}
  \begin{tabular}{@{}l cccccc@{}}
    \toprule
	   Sampling Method && IDF1 & MOTA & FP$\downarrow$ & FN$\downarrow$ & IDS$\downarrow$ \\
    \midrule
	  Center point           && 80.0 & 72.7 & 1835 & \bf 5345 & 213 \\
      Center area            && 79.8 & 72.9 & \bf 1458 & 5675 & 211\\
    \midrule
	  \textbf{LVS}   && \bf 81.5 & \bf 73.8 & 1524 & 5393 & \bf 183 \\
    \bottomrule
  \end{tabular}
    \vspace{-2mm}
  \caption{Comparisions between different sampling strategies.}
  \label{tab:sampling}
\vspace{-3mm}
\end{table}   

\vspace{1mm}
\noindent \textbf{Analysis of LVS.} In this part, we conduct an in-depth analysis on LVS. We compare it with two other keypoint sampling strategies, the center point and center area based sampling strategies, which are illustrated in Fig.~\ref{exp:lvs} (a) and (b), respectively. Similar to LVS, the center area based sampling strategy provides 9 appearance vectors for constructing the contrastive learning loss. However, the 9 keypoints are pre-defined and not adjusted according to the content of the input image.

The experimental results are reported in Tab.~\ref{tab:sampling}. According to the two primary metrics, IDF1 and MOTA, the results obtained by the center point and center area based sampling strategies are similar. This phenomenon implies that directly incorporating more appearance vectors into the training process cannot boost the performance of the model, and selecting keypoints adaptively (Fig.~\ref{exp:lvs} (c)) is important.

\begin{figure}[htbp]
\centering
\includegraphics[width=0.75\columnwidth]{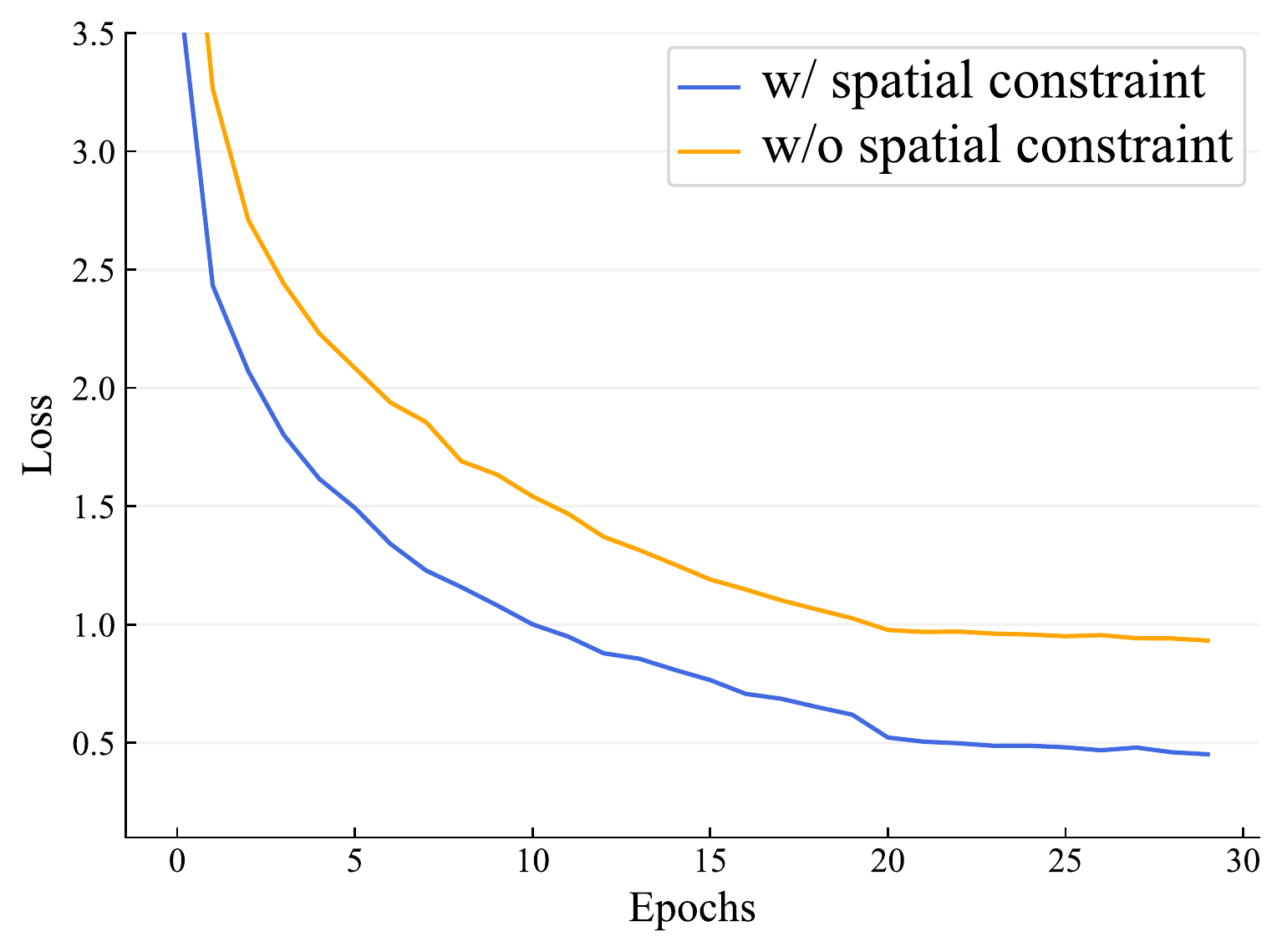} 
\caption{The loss curves of the models trained with and without the spatial constraint operator $\Phi(\cdot)$ on the MOT17 benchmark.}
\vspace{-3mm}
\label{fig:loss}
\end{figure}

\vspace{1mm}
\noindent \textbf{Analysis of the operator $\Phi(\cdot)$.} In this part, we analyze how the operator $\Phi(\cdot)$ in Eq.~(\ref{Eq2}), which restricts the selected keypoints to fall within the estimated 2D bounding box, affects the training process. The curves of the embedding head loss $L_{tcl}$ corresponding to the models trained with and without $\Phi(\cdot)$ are illustrated in Fig.~\ref{fig:loss}. We can observe that $\Phi(\cdot)$ decreases the loss value and accelerates the convergence process significantly. This observation suggests that sampling keypoints only in the estimated 2D bounding boxes is valuable because it forces the generated keypoints to reflect the target appearance information, and employing these keypoints to implement contrastive learning leads to a model with better feature extraction ability. 

\vspace{1mm}
\noindent \textbf{Analysis of the trajectory center updating strategy.} As introduced in Sec.~\ref{MTCL}, we select the hardest appearance vector of a target to update its corresponding trajectory center. To analyze whether there exists a better strategy, we compare it with other three strategies, ``Random'', ``Average'' and ``Easy''. The results are reported in Tab.~\ref{tab:updating stragety}. 

As shown in Tab.~\ref{tab:updating stragety}, ``Hard'' leads to the best results, and the performances of the other three strategies are similar. We speculate that this is because most appearance vectors of the same target generated from neighboring frames are similar, and employing them to update the trajectory center does not fully explore all information available. On the contrary, utilizing the hardest appearance vector alleviates this problem.

\begin{table}[ht]
  \centering
\small
  \setlength{\tabcolsep}{3pt}
  \begin{tabular}{@{}l ccccc@{}}
    \toprule
     Updating strategy & IDF1 & MOTA & FP$\downarrow$ & FN$\downarrow$ & IDS$\downarrow$ \\
    \midrule
    Random           & 80.3 & 73.1 & 1712 & 5382 & 190 \\
    Average            & 80.6 & 73.2 & 1669 & 5389 & 208 \\
    Easy            & 80.2 & 73.0 & 2058 & \bf5071 & 189 \\
    \midrule
    \textbf{Hard}   & \bf 81.5 & \bf 73.8 & \bf 1524 & 5393 & \bf 183 \\
    \bottomrule
  \end{tabular}
  \caption{Comparison between different trajectory center updating strategies. Random: randomly select an appearance vector; Average: Take the average of all appearance vectors; Easy: Select the appearance vector of the maximum cosine similarity with the trajectory center; Hard: Select the appearance vector with the minimum cosine similarity with the trajectory center.}
  \label{tab:updating stragety}
\vspace{-6mm}
\end{table}   

\subsection{Embedding Visualization}
\label{vis}

\begin{figure}[h]
	\centering
	\begin{subfigure}[h]{1.62in}
		\centering
		\includegraphics[width=1.8in]{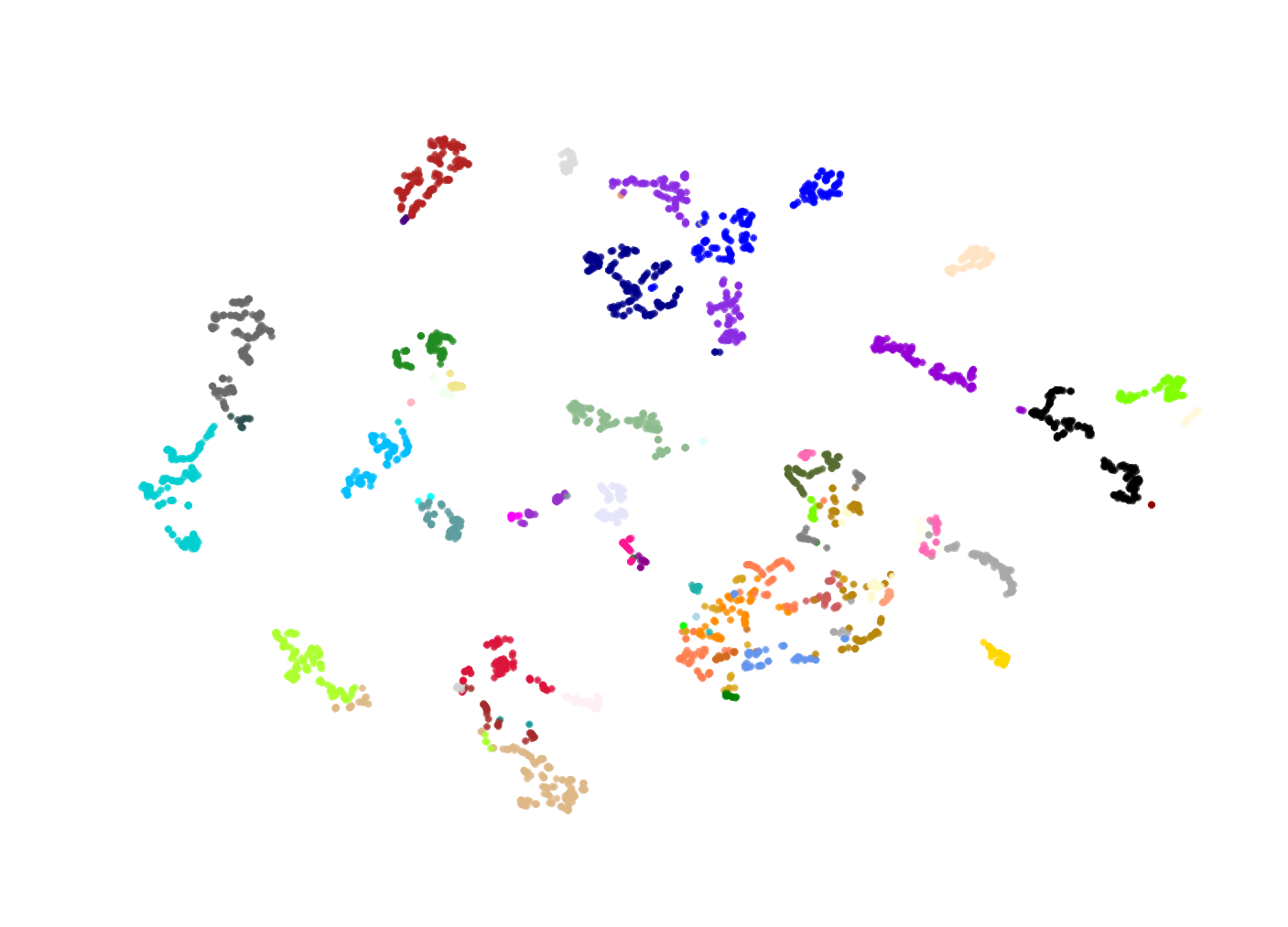}
		\caption{without MTCL}\label{tsne_Baseline}		
	\end{subfigure}
	\begin{subfigure}[h]{1.62in}
		\centering
		\includegraphics[width=1.8in]{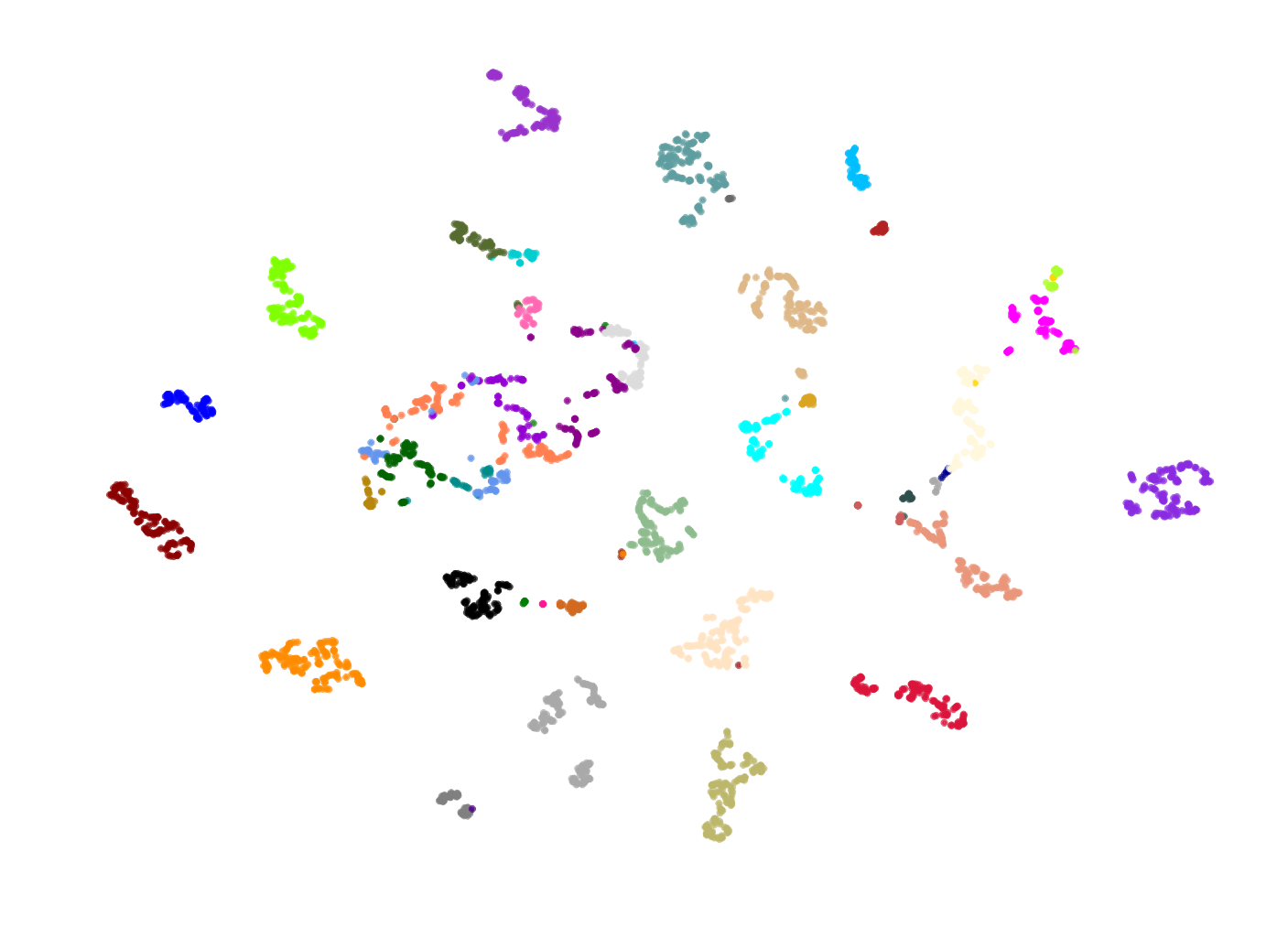}
		\caption{with MTCL}\label{tsne_MTCL}
	\end{subfigure}
	\caption{Visualizing the appearance vectors of some targets in MOT17 using the t-SNE algorithm. The points in the same color are with the same identity.}
\label{tsne}
\vspace{-2mm}
\end{figure}

In this subsection, we visualize the target appearance vectors produced by models with and without MTCL based on the t-SNE algorithm \cite{van2008visualizing}. The model with MTCL is the same as MTrack, and the model without MTCL employs the cross-entropy loss to train the embedding head of CenterNet like \cite{zhang2021fairmot}. The visualization results are given in Fig.~\ref{tsne}.

As shown in Fig.~\ref{tsne}, the representation generated by the model with MTCL is more discriminative. The vectors corresponding to the same identity are well clustered, and the ones with different identities are clearly distinguished. Hence, MTCL is effective on improving the feature extraction ability of a network.

\section{Conclusion}
In this work, we have argued that the discriminability of the extracted representation is critical for MOT. However, existing work only exploits the features in neighboring frames, and the information in the whole trajectories is ignored. To bridge this gap, we have proposed a strategy named multi-view trajectory contrastive learning, which fully exploits the intra-frame features and inter-frame features at the cost of limited computing resource. In the inference stage, a strategy named similarity-guided feature fusion has been developed to alleviate the negative influence of poor features caused by occlusion and blurring. We have verified the effectiveness of the proposed techniques on 4 public benchmarks, i.e., MOT15, MOT16, MOT17 and MOT20. The experimental results indicate that these techniques can boost the tracking performance significantly. We hope this work can serve as a new solution to producing discriminative representation in MOT.

\appendix
\section{Appendix}

In this file, we provide more details about the proposed method (MTrack) for multi-object tracking (MOT) due to the 8-pages limitation on paper length.
\section{Projection Head}

\begin{figure}[h]
\centering
\includegraphics[width=0.8\columnwidth]{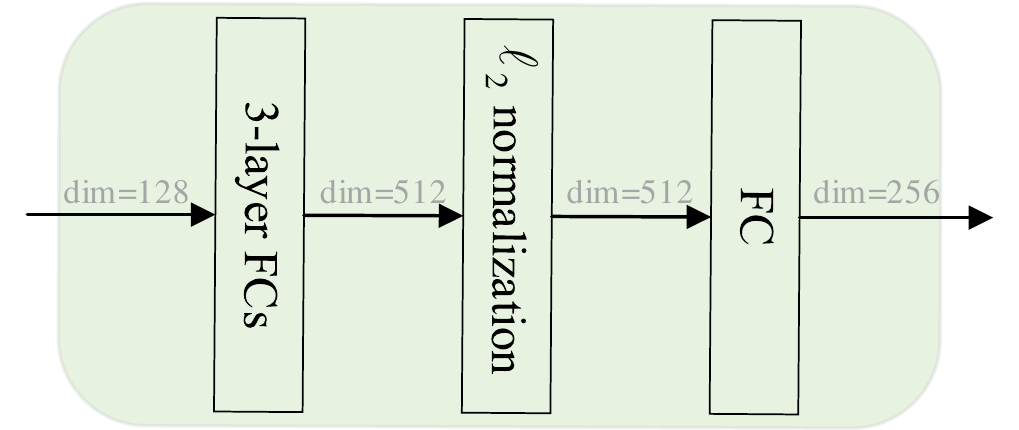} 
\caption{\textbf{Structure of the projection head.} FC: fully connected layer.}
\label{fig:proj}
\end{figure}

As shown in Fig.~3 of the paper, a projection head is added after the learnable view sampling (LVS) strategy to produce the target appearance vectors. Similar with preceding contrastive learning methods \cite{chen2020simple,caron2020unsupervised,caron2021emerging}, the projection head is composed of 4 fully connected layers (FCs). Its structure is illustrated in Fig.~\ref{fig:proj}. The output dimensions of the first 2 FCs (hidden layers) are 1024. A $\ell_{2}$ normalization operation is added after the $3_{\rm rd}$ FC.

\begin{algorithm}[htbp]
\SetAlgoLined
\DontPrintSemicolon
\SetNoFillComment
\footnotesize
\SetKwInOut{Input}{Input} \SetKwInOut{Output}{Output}
\Input{$T^{(t - 1)} = \{(\textbf{c}, \textbf{b}, \textbf{te})_j^{(t-1)}\}_{j=1}^{M}$ // $T^{(t - 1)}$ is the trajectory waiting to be matched from frame $t-1$. \textbf{c} and \textbf{te} are the 2D center and embedding vector of the target in $T^{(t - 1)}$. $\textbf{b} = (w, h)$ is the box size. \\
$\hat{D}^{(t)} = \{(\hat{\textbf c}, \hat{\textbf b}, \textbf {e})_i^{(t)}\}_{i=1}^{N}$ // $\hat{D}^{(t)}$ are the detected\\ candidates in frame $t$. $\hat{\textbf c}$, $\hat{\textbf b}$, and $\textbf {e}$ are their centers, box \\sizes, and embedding vectors, respectively. \\ 
Trajectory memory sequence length $Q$; \\ 
Association threshold $\kappa_1$,$\kappa_2$ and $\kappa_3$;}
\Output{$T^{(t)} = \{(\textbf{p}, \textbf{s}, \textbf{te})_i^{(t)}\}_{i=1}^{N}$, Updated trajectory in the current frame;}

\textbf{Initialization:} \; 
  $L \leftarrow length\ \ of \ \ video$ \;
  $T^{(t)} \leftarrow \emptyset$ \;
  $UT^{(t)} \leftarrow \emptyset$ \ \ // unactived trajectory \;
  $M \leftarrow \emptyset$ \ \ // trajectory memory bank \;
  $W \leftarrow Cost(D^{(t)}, T^{(t-1)})$ \;

\For{$t \leftarrow 1 \ to \ L$}{
  \BlankLine
  \BlankLine
  \tcc{Step 1: association with appearance similarity}
  $W_{cos}=Cosine \ \ distance(\textbf{e}^{(t)}, \textbf{te}^{(t-1)})$ \;
  $mt_1, ut_1, ud_1 \leftarrow Hungarian(KF_{t}(W_{cos}), \kappa_1)$ \;
  \BlankLine
  \BlankLine
  \tcc{Step2: association with IoU}
  $W_{IoU} \leftarrow IoU \ \ distance((\hat{\textbf c}, \hat{\textbf b})_{ud_1}^{(t)}, (\textbf{c}, \textbf{b})_{ut_1}^{(t-1)})$ \;
  $mt_2, ut_2, ud_2 \leftarrow Hungarian(W_{IoU}, \kappa_2)$ \;
  \For{$j \ \ in \ \ ut_2$}{
        \If {$missing \ \ frame \ \ length \ \ > \ \ \lambda$} {
        delete \ \ $T_{j}$
        }
      }
  \BlankLine
  \BlankLine
  \tcc{Step3: associate unactivated trajectory with unmatched detections}
  $W_{IoU2} \leftarrow IoU \ \ distance((\hat{\textbf c}, \hat{\textbf b})_{ud_2}^{(t)}, UT^{(t-1)})$ \;
  $mt_3, ut_3, ud_3 \leftarrow Hungarian(W_{IoU2}, \kappa_3)$ \;
  \For{$j \ \ in \ \ ut_3$}{
      delete \ \ $UT_{j}$
    }
  \BlankLine
  \BlankLine
  \tcc{Update matched trajectory}
  \For{$i \ \ in \ \ (mt_1 \cup mt_2 \cup mt_3)$}{
  $T^{(t)} \leftarrow T^{(t)} \cup (\hat{\textbf{c}}_i^{(t)}, \hat{\textbf{b}}_i^{(t)}, \hat{\textbf{e}}_i^{(t)})$ \;
  update $M_{i}$ and $tc_{i}^{(t)}$ with $\beta^{t}$ by SGFF
  }
  \BlankLine
  \BlankLine
  \tcc{Initialize new trajectory}
  \For{$ui \ \ in \ \ ud_3$}{
        $UT^{(t)} \leftarrow T^{(t)} \cup (\hat{\textbf{c}}_{ui}^{(t)}, \hat{\textbf{b}}_{ui}^{(t)}, \hat{\textbf{e}}_{ui}^{(t)})$ \
     }
}
Return: $T^{(t)}$
\caption{Pseudo-code of Tracking.}
\label{alg:association}
\end{algorithm}

\section{Tracking Algorithm}
\label{sec:tracking}

We detail the association algorithm of MTrack in this section. The algorithm associates targets to trajectories mainly based on the appearance similarity, and the motion information is used as auxiliary clues to boost the association accuracy. The pseudo code of this association algorithm is presented in Algorithm~\ref{alg:association}.

The association algorithm mainly comprises three steps. In Step 1 (line 8 in Algorithm~\ref{alg:association}), we associate targets to trajectories based on the computed appearance similarities. Notably, we only calculate the similarity values between the targets and their surrounding trajectories. The trajectories that are far away from the targets are not considered. In Step 2 (line 10--16 in Algorithm~\ref{alg:association}), the unmatched targets are associated with the unmatched trajectories that are not newly generated in the last frame (activated trajectories) based on the predicted motion information (predicted by the Kalman model \cite{welch1995introduction}). In Step 3 (line 17--21 in Algorithm~\ref{alg:association}), we link the still unmatched targets to the remaining trajectories which are newly generated in the last frame (unactivated trajectories). 

If a detected target is still not matched with any trajectory, a new trajectory is established. In addition, for matched trajectories, we use our similarity-guided feature fusion (SGFF) strategy to update the trajectory embeddings. If a trajectory is not matched with any target for $\lambda$ frames, this trajectory will not be considered in the following association process. $\lambda$ is set as 15 in our implementation. Besides, $\kappa_1$, $\kappa_2$, and $\kappa_3$ are there threshold parameters in the Hungarian algorithm \cite{kuhn1955hungarian}. They are set as $0.3$, $0.5$, and $0.7$, respectively.

We have validated our method on the MOT15 \cite{leal2015motchallenge}, MOT16 \cite{milan2016mot16}, MOT17 \cite{milan2016mot16} and MOT20 \cite{dendorfer2020mot20} benchmarks. Since the targets in MOT16 and MOT17 are not crowded and move regularly, we adopt the trajectory filling strategy proposed in MAT \cite{han2020mat} as a post-processing operation to reduce the false negative samples (FN) caused by trajectory disconnection. For MOT20, the targets are very crowded and occluded seriously. Therefore, we adjust the thresholds to associate the targets with a stricter rule ($\kappa_1 = 0.25$ and $\kappa_3 = 0.5$).

\section{Additional Experiments}

The effectiveness of various components in MTrack has been confirmed by the extensive experiments in the paper. In this section, we present some additional experimental results to further analyze our method.

\subsection{Analysis of the keypoint sampling number $N_{k}$} 

\begin{figure}[t]
  \centering
  \begin{subfigure}[t]{3.1in}
    \centering
    \includegraphics[width=3.1in]{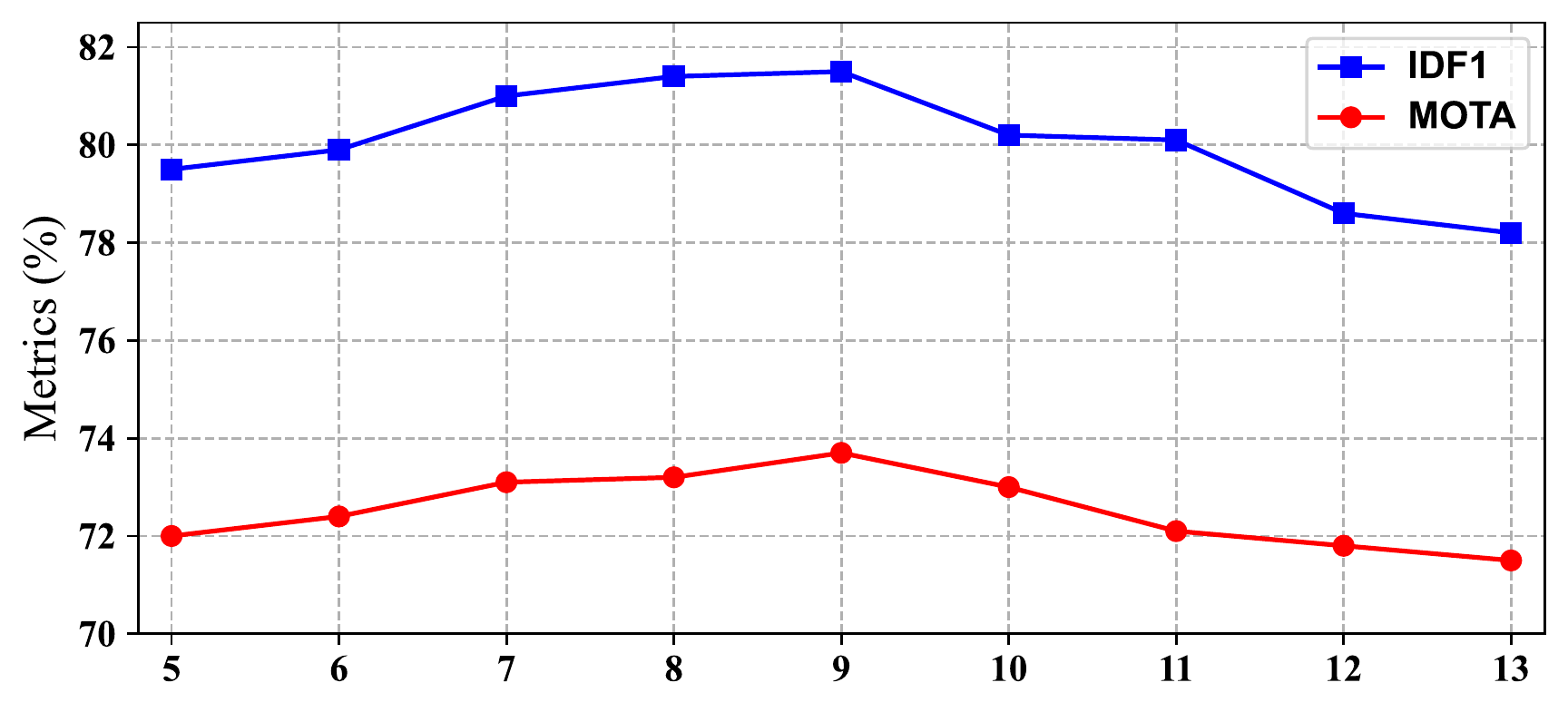}
    \subcaption*{Sampling number $N_{k}$}
  \end{subfigure}
  \caption{\textbf{Analysis of the keypoint sampling number $N_{k}$ in LVS on the tracking performance.} The experiment is conducted on our divided MOT17 dataset.}
\label{fig:sample}
\vspace{-3mm}
\end{figure}

As introduced in Sec.3.2 of the paper, LVS selects multiple keypoints adaptively for every target to produce appearance vectors, and the number of selected keypoints is denoted as $N_{k}$. In this part, we study how $N_{k}$ affects the tracking performance. The results are reported in Fig.~\ref{fig:sample}.

As shown in Fig.~\ref{fig:sample}, selecting too many and too few keypoints both lead to unsatisfactory results. The best performance is obtained when $N_{k}$ is 9. We speculate that when $N_{k}$ is small, insufficient features are extracted. If $N_{k}$ is large, predicting all the offsets of key samples requires a larger network that is hard to be optimized.

\subsection{Analysis of the momentum parameter $\alpha$} 

\begin{figure}[htbp]
  \centering
  \begin{subfigure}[htbp]{2.3in}
    \centering
    \includegraphics[width=2.3in]{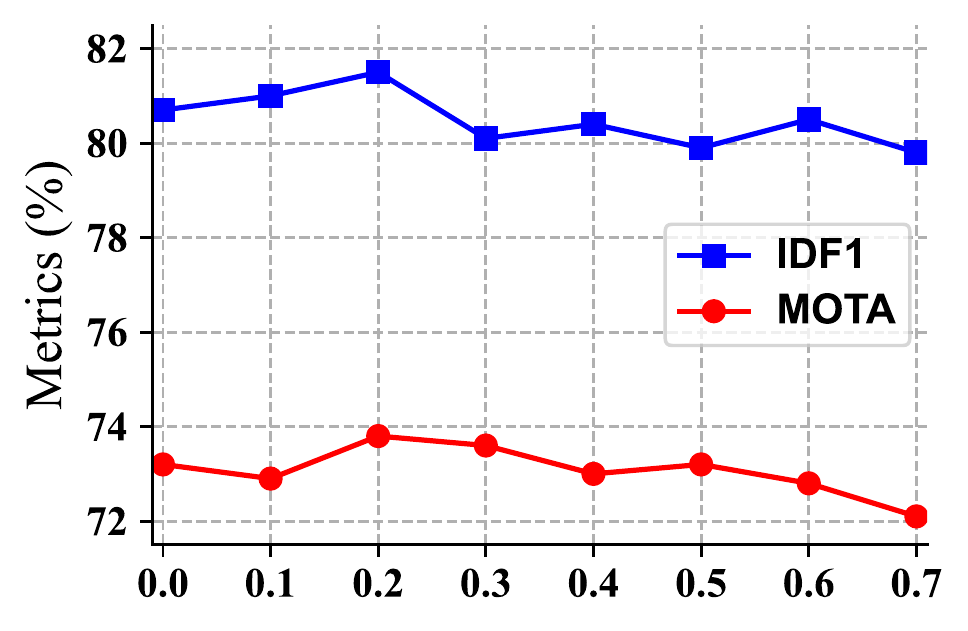}
    \subcaption*{Momentum value $\alpha$}
  \end{subfigure}
  \caption{The impact of the momentum value $\alpha$ on devided MOT17 datasets.}\label{momentum}
\label{exp:momentum}
\vspace{-3mm}
\end{figure}

The momentum update strategy is an important component in existing contrastive learning algorithms \cite{he2020momentum, chen2020improved, caron2021emerging}. In this work, we use the momentum update strategy to update the \textit{trajectory centers} stored in our trajectory-center memory bank (TMB). As explained in the paper, this process is controlled by the hyper-parameter $\alpha$. In this part, we study how setting $\alpha$ to different values affects the tracking performance. The experiment is conducted on the divided MOT17 dataset, and the results are illustrated in Fig.~\ref{exp:momentum}.

It can be observed from Fig.~\ref{exp:momentum} that $\alpha$ of a small value (less than 0.2) leads to better performance on both IDF1 and MOTA, which means the trajectory centers should be updated by the produced appearance vectors quickly. This observation confirms that the appearance vectors are of good quality to some extent. According to the results in Fig.~\ref{exp:momentum}, we set $\alpha$ to 0.2.

\subsection{Analysis of the trajectory memory length $Q$} 
$Q$ denotes the maximum length of the trajectory memory, which is used in the proposed similarity-guided feature fusion (SGFF) strategy. In this part, we study how $Q$ influences the tracking precision of trained models. The experiment is conducted on the divided MOT17 benchmark, and $Q$ is set as 10, 20, 30, 40, 50, and 60, respectively. The results are reported in Tab.~\ref{tab:sequence}.

From Tab.~\ref{tab:sequence}, we can observe that $Q$ has a minor influence on the metric MOTA, but affects the metric IDF1 significantly. When $Q$ is too small or large, the performance of the obtained tracker is not satisfying, and the best performance is obtained when $Q$ is 30. We speculate that this is because when $Q$ is small, the maintained vectors are not very informative. Meanwhile, when $Q$ is large, too old vectors are employed to produce the current trajectory representation, which is harmful. In our implementation, $Q$ is set to 30.

\begin{table}[htbp]
  \centering
\small
  \setlength{\tabcolsep}{3pt}
  \begin{tabular}{@{}c ccccc@{}}
    \toprule
     length Q & IDF1 & MOTA & FP$\downarrow$ & FN$\downarrow$ & IDS$\downarrow$ \\
    \midrule
    10            & 79.4 & 73.8 & \bf 1479 & 5434 & 183\\
      20            & 81.0 & 73.8 & 1504 & 5408 & \bf 177\\
      30            & \bf 81.5 & \bf 73.8 & 1524 & 5393 & 183\\
      40            & 81.4 & 73.8 & 1525 & 5384 & 195\\
      50            & 81.2 & 73.7 & 1527 & \bf 5382 & 200\\
      60            & 80.8 & 73.7 & 1529 & 5378 & 203\\
    \bottomrule
  \end{tabular}
    \vspace{-2mm}
  \caption{Comparision between different maximum lengths of trajectory memory.}
  \label{tab:sequence}
\end{table} 

\section{Additional Visualization}

\subsection{Qualitative results}

\begin{figure*}[t]
\centering
\includegraphics[width=1.9\columnwidth]{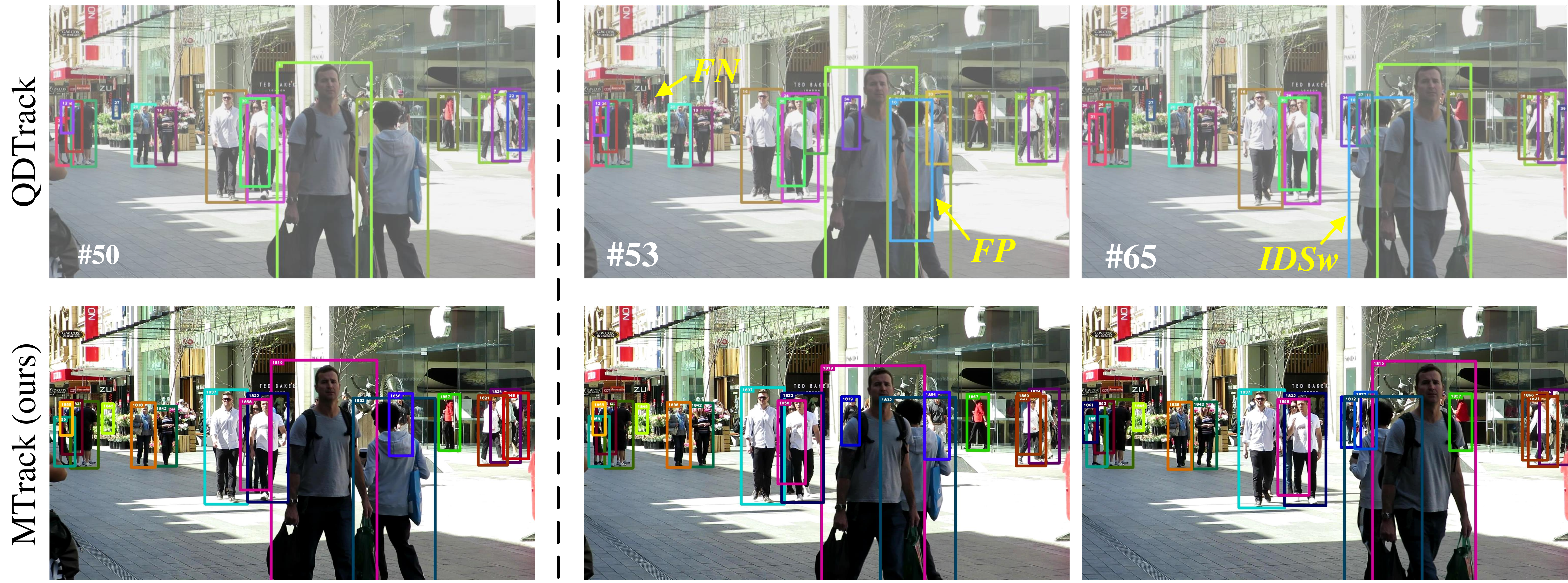} 
\caption{\textbf{Qualitative comparison between MTrack and QDTrack}. They are compared on a challenging case selected from the testing split of MOT17. FP, FN and IDSw represent different kinds of error types, which are false positive, false negative and identity switch, respectively.}
\label{fig:quality}
\end{figure*}

As introduced in the paper, our proposed MTCL provides meaningful and rich view representation for contrastive learning, which contributes to improving the discriminability of learned representation. To show the effectiveness of our approach, we select a challenging case from MOT to compare the performance of MTrack with that of QDTrack \cite{pang2021quasi}, a similar tracker adopting the RPN network \cite{ren2015faster} to generate dense samples and conduct contrastive learning. The results are illustrated in Fig.~\ref{fig:quality}. As shown, QDTrack fails to track the targets when they are occluded or small. On the contrary, MTrack tracks the targets correctly in this challenging case. We infer this is because MTrack can produce more informative representation due to MTCL.

\subsection{Visualization of tracking results}

\begin{figure*}[htbp]
\centering
\includegraphics[width=2.1\columnwidth]{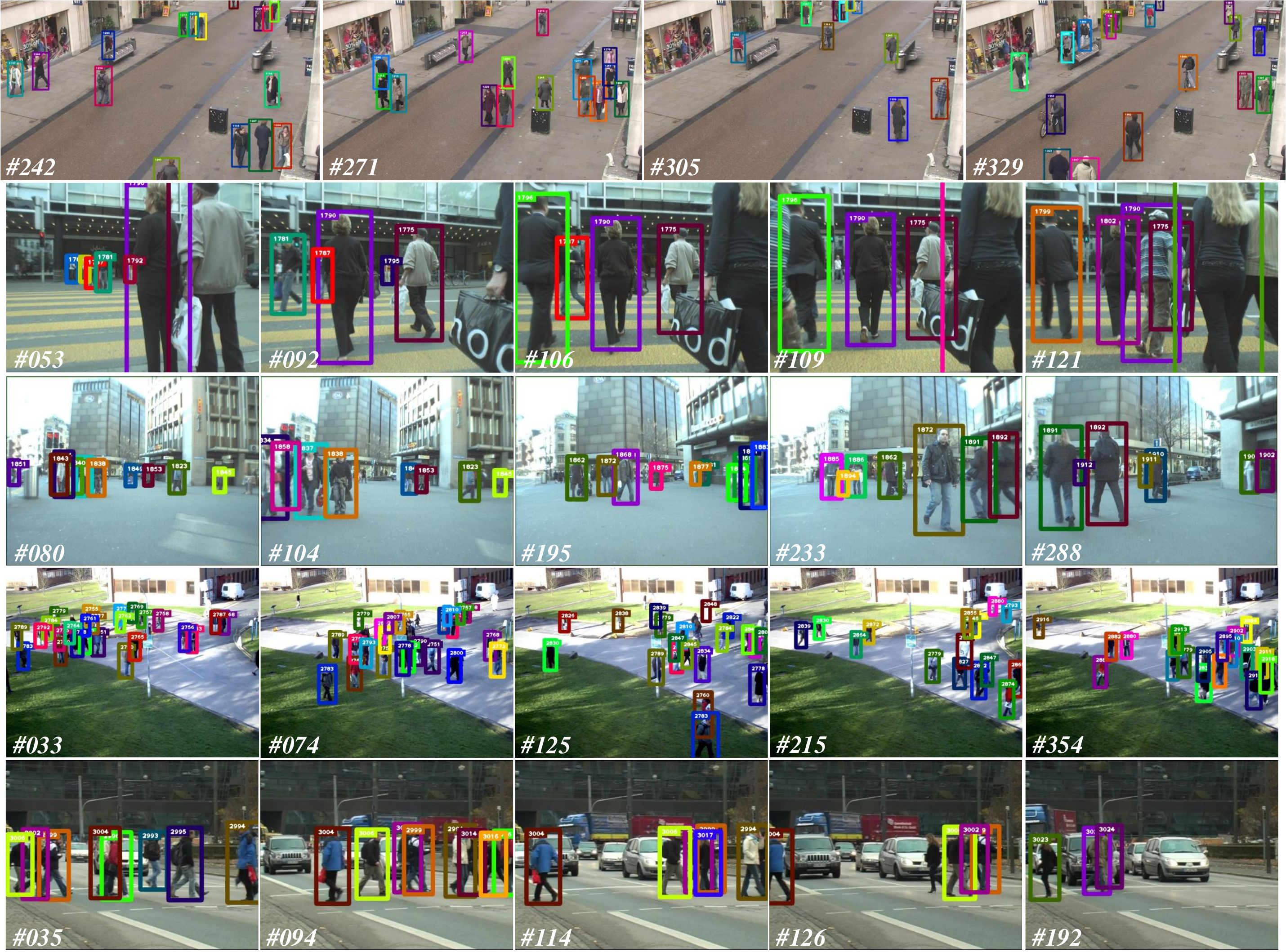} 
\caption{Tracking results of MTrack on the testing split of MOT15.}
\label{fig:mot15}
\end{figure*}

\begin{figure*}[htbp]
\centering
\includegraphics[width=2.1\columnwidth]{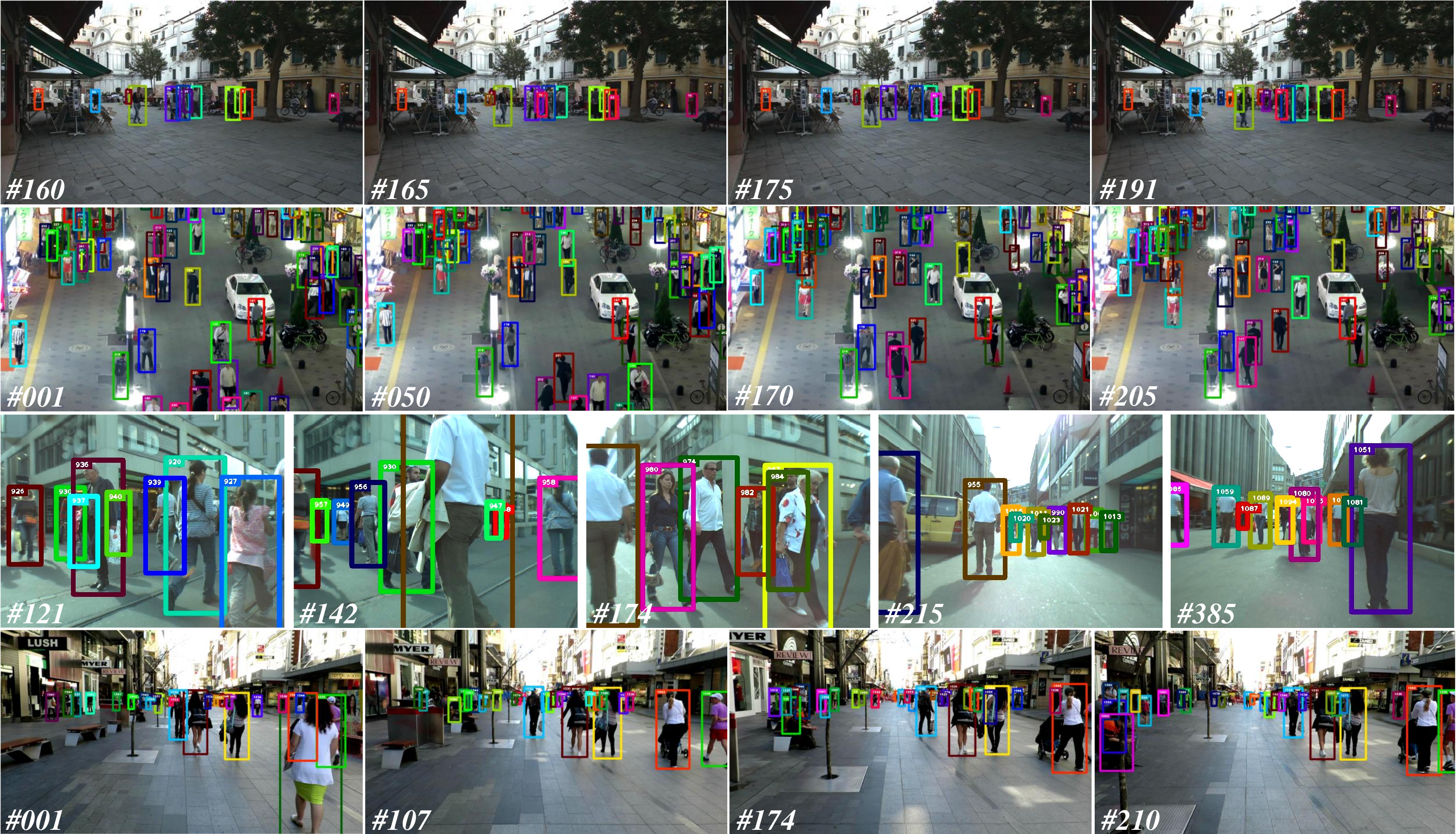} 
\caption{Tracking results of MTrack on the testing split of MOT17.}
\label{fig:mot17}
\end{figure*}

\begin{figure*}[htbp]
\centering
\includegraphics[width=2.1\columnwidth]{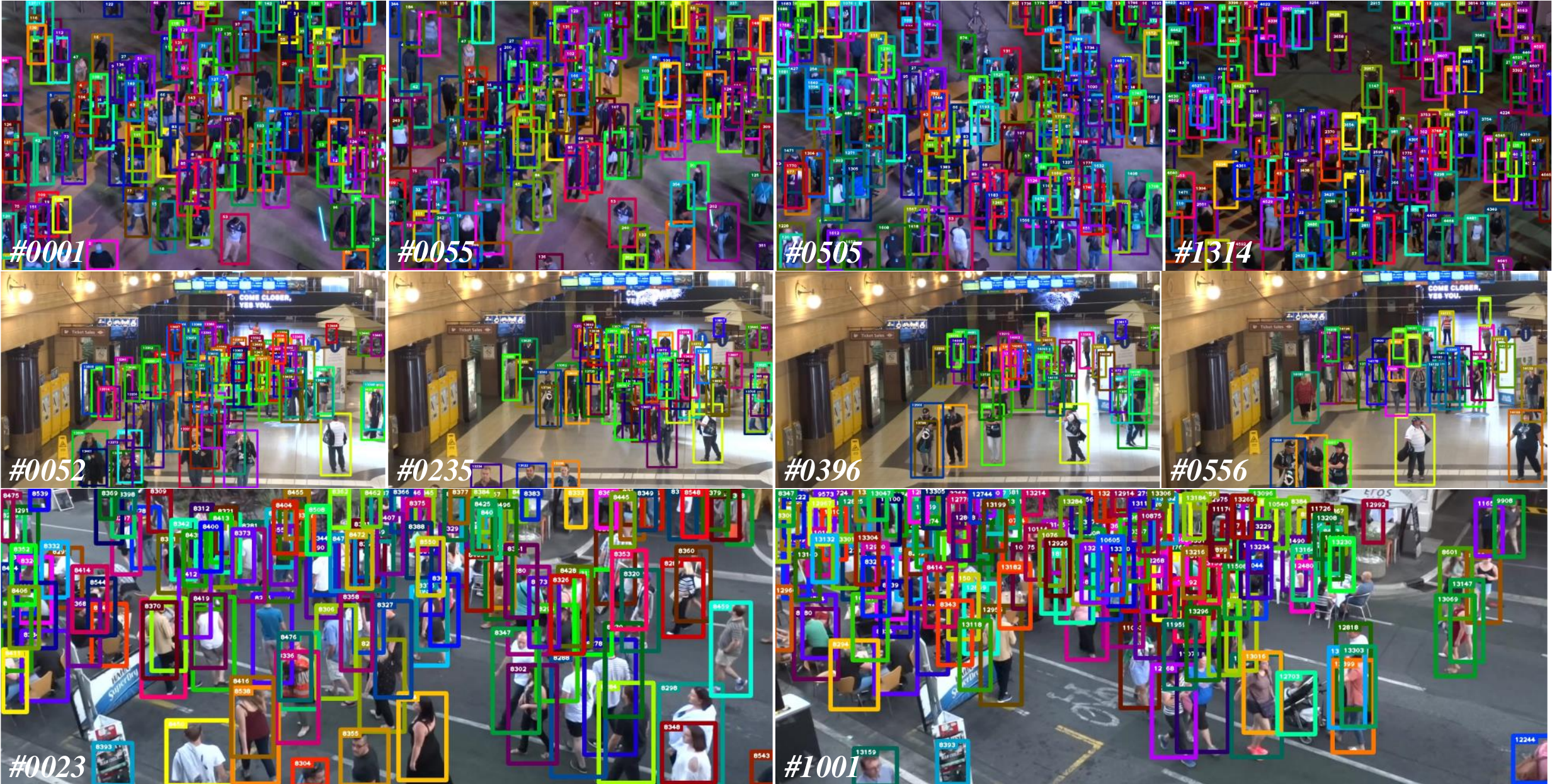} 
\caption{Tracking results of MTrack on the testing split of MOT20.}
\label{fig:mot20}
\end{figure*}

More challenging cases are visualized and presented in Fig.~\ref{fig:mot15}, Fig.~\ref{fig:mot17}, and Fig.~\ref{fig:mot20} to further confirm the superiority of MTrack. The shown cases are chosen from the test splits of MOT15 \cite{leal2015motchallenge}, MOT17 \cite{milan2016mot16} and MOT20 \cite{dendorfer2020mot20}. They cover various situations, including indoor and outdoor scenarios, light and dark brightness, huge and small targets, etc.

{\small
\bibliographystyle{ieee_fullname}
\bibliography{reference}
}

\clearpage
\end{document}